# Answer Set Planning Under Action Costs

**Thomas Eiter** EITER@KR.TUWIEN.AC.AT
**Wolfgang Faber** FABER@KR.TUWIEN.AC.AT
*Institut für Informationssysteme, TU Wien*
*Favoritenstr. 9-11, A-1040 Wien, Austria*

**Nicola Leone** LEONE@UNICAL.IT
*Department of Mathematics, University of Calabria*
*I-87030 Rende (CS), Italy*

**Gerald Pfeifer** PFEIFER@DBAI.TUWIEN.AC.AT
**Axel Polleres** POLLERES@KR.TUWIEN.AC.AT
*Institut für Informationssysteme, TU Wien*
*Favoritenstr. 9-11, A-1040 Wien, Austria*

## Abstract

Recently, planning based on answer set programming has been proposed as an approach towards realizing declarative planning systems. In this paper, we present the language $\mathcal{K}^c$, which extends the declarative planning language $\mathcal{K}$ by action costs. $\mathcal{K}^c$ provides the notion of admissible and optimal plans, which are plans whose overall action costs are within a given limit resp. minimum over all plans (i.e., cheapest plans). As we demonstrate, this novel language allows for expressing some nontrivial planning tasks in a declarative way. Furthermore, it can be utilized for representing planning problems under other optimality criteria, such as computing "shortest" plans (with the least number of steps), and refinement combinations of cheapest and fastest plans. We study complexity aspects of the language $\mathcal{K}^c$ and provide a transformation to logic programs, such that planning problems are solved via answer set programming. Furthermore, we report experimental results on selected problems. Our experience is encouraging that answer set planning may be a valuable approach to expressive planning systems in which intricate planning problems can be naturally specified and solved.

## 1. Introduction

Recently, several declarative planning languages and formalisms have been introduced, which allow for an intuitive encoding of complex planning problems involving ramifications, incomplete information, non-deterministic action effects, or parallel actions (see e.g., Giunchiglia & Lifschitz, 1998; Lifschitz, 1999b; Lifschitz & Turner, 1999; McCain & Turner, 1998; Giunchiglia, 2000; Cimatti & Roveri, 2000; Eiter et al., 2000b, 2003b).

While these systems are designed to generate any plans that accomplish the planning goals, in practice one is often interested in particular plans that are optimal with respect to some objective function by which the quality (or the cost) of a plan is measured. A common and simple objective function is the length of the plan, i.e., the number of time steps to achieve the goal. Many systems are tailored to compute shortest plans. For example, CMBP (Cimatti & Roveri, 2000) and GPT (Bonet & Geffner, 2000) compute shortest plans in which each step consists of a single action, while the Graphplan algorithm (Blum & Furst, 1997) and descendants (Smith & Weld, 1998; Weld,

Anderson, & Smith, 1998) compute shortest plans where in each step actions might be executed in parallel.

However, there are other, equally important objective functions to consider. In particular, if executing actions causes some cost, we may desire a plan which minimizes the overall cost of the actions.

In answer set planning (Subrahmanian & Zaniolo, 1995; Dimopoulos, Nebel, & Koehler, 1997; Niemelä, 1998; Lifschitz, 1999b), a recent declarative approach to planning where plans are encoded by the answer sets of a logic program, the issue of optimal plans under an objective value function has not been addressed in detail so far (see Section 8 for more details). In this paper, we address this issue and present an extension of the planning language $\mathcal{K}$ (Eiter et al., 2000b, 2003b), where the user may associate costs with actions, which are then taken into account in the planning process. The main contributions of our work are as follows.

- We define syntax and semantics of the planning language $\mathcal{K}^c$, which modularly extends the language $\mathcal{K}$: Costs are associated to an action by extending the action declarations with an optional cost construct which describes the cost of executing the respective action.

  The action costs can be static or dynamic, as they may depend on the current stage of the plan when an action is considered for execution. Dynamic action costs are important and have natural applications, as we show on a simple variant of the well-known Traveling Salesperson Problem, which is cumbersome to model and solve in other, similar languages.

- We analyze the computational complexity of planning in the language $\mathcal{K}^c$, and provide completeness results for major planning tasks in the propositional setting, which locate them in suitable slots of the Polynomial Hierarchy and in classes derived from it. These results provide insight into the intrinsic computational difficulties of the respective planning problems, and give a handle for efficient transformations from optimal planning to knowledge representation formalisms, in particular to logic programs.

- We show, in awareness of the results of the complexity analysis, how planning with action costs can be implemented by a transformation to answer set programming, as done in a system prototype that we have developed. The prototype, ready for experiments, is available at `http://www.dlvsystem.com/K/`.

- Finally, we present some applications which show that our extended language is capable of easily modeling optimal planning under various criteria: computing (1) "cheapest" plans (which minimize overall action costs); (2) "shortest" plans (with the least number of steps); and, refinement combinations of these, viz. (3) shortest plans among the cheapest, and (4) cheapest plans among the shortest. Notice that, to our knowledge, task (3) has not been addressed in other works so far.

The extension of $\mathcal{K}$ by action costs provides a flexible and expressive tool for representing various problems. Moreover, since $\mathcal{K}$'s semantics builds on states of knowledge rather than on states of the world, we can deal with both incomplete knowledge *and* plan quality, which is, to the best of our knowledge, completely novel.

Our experience is encouraging that answer set planning, based on powerful logic programming engines, allows for the development of declarative planning systems in which intricate planning tasks can be specified and solved. This work complements and extends the preliminary results presented in our previous work (Eiter et al., 2002a).

The remainder of this paper is organized as follows. In the next section, we briefly review the language $\mathcal{K}$ by informally presenting its main constituents and features on a simple planning example. After that, we define in Section 3 the extension of $\mathcal{K}$ by action costs, and consider some first examples for the usage of $\mathcal{K}^c$. Section 4 is devoted to the analysis of complexity issues. In Section 5, we consider applications of $\mathcal{K}^c$. We show that various types of particular optimization problems can be expressed in $\mathcal{K}^c$, and also consider some practical examples. In Section 6, we present a transformation of $\mathcal{K}^c$ into answer set programming, and in Section 7, we report about a prototype implementation and experiments. After a discussion of related work in Section 8, we conclude the paper with an outlook on ongoing and future work.

## 2. Short Review of Language $\mathcal{K}$

In this section, we give a brief informal overview of the language $\mathcal{K}$, and refer to (Eiter et al., 2003b) and to the Appendix for formal details. We assume that the reader is familiar with the basic ideas of planning and action languages, in particular with the notions of actions, fluents, goals and plans. For illustration, we shall use the following planning problem as a running example.

**Problem 1 [Bridge Crossing Problem]** *Four persons want to cross a river at night over a plank bridge, which can only hold up to two persons at a time. They have a lamp, which must be used when crossing. As it is pitch-dark and some planks are missing, someone must bring the lamp back to the others; no tricks (like throwing the lamp or halfway crosses, etc.) are allowed.*

**Fluents and states.** A state in $\mathcal{K}$ is characterized by the truth values of fluents, describing relevant properties of the domain of discourse. A fluent may be true, false, or unknown in a state – that is, states in $\mathcal{K}$ are *states of knowledge*, as opposed to states of the world where each fluent is either true or false (which can be easily enforced in $\mathcal{K}$, if desired). Formally, a *state* is any consistent set $s$ of (possibly negated) legal fluent instances.

An action is applicable only if some precondition (a list of literals over some fluents) holds in the current state. Its execution may cause a modification of truth values of some fluents.

**Background knowledge.** Static knowledge which is invariant over time in a $\mathcal{K}$ planning domain is specified in a normal (disjunction-free) Datalog program $\Pi$ that has a single answer set and can be viewed as a set of facts. For our example, the background knowledge specifies the four persons:

```
person(joe).  person(jack).  person(william).  person(averell).
```

**Type declarations.** Each fluent or action must have a declaration where the ranges of its arguments are specified. For instance,

```
crossTogether(X, Y) requires person(X), person(Y), X < Y.
```
[1]

specifies the arguments of the action `crossTogether`, where two persons cross the bridge together, while

```
across(X) requires person(X).
```

1. "<" here is used instead of inequality to avoid symmetric rules.

specifies a fluent describing that a specific person is on the other side of the river. Here the literals after "requires" must be classical literals of the static background knowledge (like person(X) and person(Y)), or literals of built-in predicates (such as $X < Y$). Our implementation of $\mathcal{K}$, the DLV$^{\mathcal{K}}$-system (Eiter, Faber, Leone, Pfeifer, & Polleres, 2003a), currently supports the built-in predicates "$A < B$", "$A <= B$", "$A != B$" with the obvious meaning of less-than, less-or-equal and inequality for strings and numbers, the arithmetic built-ins "$A = B + C$" and "$A = B * C$" which stand for integer addition and multiplication, and the predicate "#int(X)" which enumerates all integers (up to a user-defined limit).

**Causation rules.** Causation rules ("rules" for brevity) are syntactically similar to rules of the action language $\mathcal{C}$ (Giunchiglia & Lifschitz, 1998; Lifschitz, 1999a; Lifschitz & Turner, 1999) and are of the basic form:

```
caused f if B after A.
```

where A is a conjunction of fluent and action literals, possibly including default negation, B is a conjunction of fluent literals, again possibly including default negation, and f is a fluent literal. Informally, such a rule reads: if B is known to be true in the current state and A is known to be true in the previous state, then f is known to be true in the current state as well. Both the if-part and the after-part are allowed to be empty (which means that they are true). A causation rule is called *dynamic*, if its after-part is not empty, and is called *static* otherwise.

Causation rules are used to express effects of actions or ramifications. For example,

```
caused across(X) after cross(X), -across(X).
caused -across(X) after cross(X), across(X).
```

describe the effects of a single person crossing the bridge in either direction.

**Initial state constraints.** Static rules can apply to all states or only to the initial states (which may not be unique). This is expressed by the keywords "always :" and "initially :" preceding sequences of rules where the latter describes *initial state constraints* that must be satisfied only in the initial state. For example,

```
initially : caused -across(X).
```

enforces the fluent across to be false in the initial state for any X satisfying the declaration of the fluent across, i.e., for all persons. The rule is irrelevant for all subsequent states.

**Executability of actions.** This is expressed in $\mathcal{K}$ explicitly. For instance,

```
executable crossTogether(X, Y) if hasLamp(X).
executable crossTogether(X, Y) if hasLamp(Y).
```

declares that two persons can jointly cross the bridge if one of them has a lamp. The same action may have multiple executability statements. A statement

```
executable cross(X).
```

with empty body says that cross is always executable, provided that the type restrictions on X are respected. Dually,

```
nonexecutable a if B.
```

prohibits the execution of action a if condition B is satisfied. For example,

```
nonexecutable crossTogether(X, Y) if differentSides(X, Y).
```

says that persons X and Y can not cross the bridge together if they are on different sides of the bridge. In case of conflicts, `nonexecutable A` overrides `executable A`.

**Default and strong negation.** $\mathcal{K}$ supports strong negation ("¬," also written as "-"). Note, however, that for a fluent `f`, in a state neither `f` nor `-f` needs to hold. In this case the knowledge about `f` is incomplete. In addition, weak negation ("`not`"), interpreted like default negation in answer set semantics (Gelfond & Lifschitz, 1991), is permitted in rule bodies. This allows for natural modeling of inertia and default properties, as well as dealing with incomplete knowledge in general. For example,

```
caused hasLamp(joe) if not hasLamp(jack), not hasLamp(william), not hasLamp(averell).
```

expresses the conclusion that by default, `joe` has the lamp, whenever it is not evident that any of the other persons has it.

**Macros.** $\mathcal{K}$ provides a number of macros as "syntactic sugar". For example,

```
inertial across(X).
```

informally states that `across(X)` holds in the current state, if `across(X)` held at the previous state, unless `-across(X)` is explicitly known to hold. This macro expands to the rule

```
caused across(X) if not -across(X) after across(X).
```

Moreover, we can "totalize" the knowledge of a fluent by declaring `total f.` which is a shortcut for

```
caused f if not -f.     caused -f if not f.
```

The intuitive meaning of these rules is that unless a truth value for `f` can be derived, the cases where `f` resp. `-f` is true will both be considered.

**Planning domains and problems.** In $\mathcal{K}$, a *planning domain* $PD = \langle \Pi, \langle D, R \rangle \rangle$ has a background knowledge $\Pi$, action and fluent declarations $D$, and rules and executability conditions $R$; a *planning problem* $\mathcal{P} = \langle PD, q \rangle$ has a planning domain $PD$ and a *query*

$$q = g_1, \ldots, g_m, \texttt{not}\ g_{m+1}, \ldots, \texttt{not}\ g_n \ ?\ (l)$$

where $g_1, \ldots, g_n$ are ground fluents and $l \geq 0$ is the plan length. For instance, the goal query

```
across(joe), across(jack), across(william), across(averell)? (5)
```

asks for plans which bring all four persons across in 5 steps.

Plans are defined using a transition-based semantics, where the execution of a set of actions transforms a current state into a new state. An *(optimistic) plan* for $\mathcal{P}$ is a sequence $P = \langle A_1, \ldots, A_l \rangle$ of sets of action instances $A_1, A_2, \ldots, A_l$ in a *trajectory* $T = \langle \langle s_0, A_1, s_1 \rangle, \langle s_1, A_2, s_2 \rangle, \ldots, \langle s_{l-1}, A_l, s_l \rangle \rangle$ from a legal initial state $s_0$ to state $s_l$ in which all literals of the goal are true. That is, starting in $s_0$, the legal transition $t_1 = \langle s_0, A_1, s_1 \rangle$, modeling the execution of the actions in $A_1$ (which must be executable), transforms $s_0$ into the state $s_1$. This is then followed by legal transitions $t_i = \langle s_{i-1}, A_i, s_i \rangle$, for $i = 2, 3, \ldots, l$ (cf. Appendix for details). A plan is *sequential*, if $|A_i| \leq 1$ for all $i = 1, \ldots, l$, i.e., each step consists of at most one action; such plans can be enforced by including the keyword `noConcurrency`.

Besides optimistic plans, in $\mathcal{K}$ we also support stronger *secure (or conformant) plans*. A secure plan must be guaranteed to work out under all circumstances (Eiter et al., 2000b), regardless of incomplete information about the initial state and possible nondeterminism in the action effects.

For better readability, in the following we will not always describe $\mathcal{K}$ planning problems $\mathcal{P}$ strictly in terms of sets of declarations, rules and executability conditions, but optionally use the more compact representation of $\mathcal{K}$ *programs* of the following general form:

$$\begin{array}{ll} \texttt{fluents:} & F_D \\ \texttt{actions:} & A_D \\ \texttt{initially:} & I_R \\ \texttt{always:} & C_R \\ \texttt{goal:} & q \end{array}$$

where the (optional) sections `fluents` through `always` consist of lists of fluent declarations $F_D$, action declarations $A_D$, initial state constraints $I_R$ and executability conditions and causation rules $C_R$, respectively. Together with the background knowledge $\Pi$ and the goal query $q$, they specify a $\mathcal{K}$ planning problem $\mathcal{P} = \langle\langle\Pi, \langle D, R\rangle\rangle, q\rangle$, where $D$ is given by $F_D$ plus $A_D$ and $R$ by $I_R$ plus $C_R$. [2]

### 2.1 Solving the Bridge Crossing Problem

Using the above constructs, a $\mathcal{K}$ encoding of the Bridge Crossing Problem, assuming that `joe` initially carries the lamp, is shown in Figure 1. There are simple five-step plans ($l = 5$), in which `joe` always carries the lamp and brings all others across. One of them is:

$$P = \langle\ \{\texttt{crossTogether(joe, jack)}\},\ \{\texttt{cross(joe)}\},\ \{\texttt{crossTogether(joe, william)}\},\ \{\texttt{cross(joe)}\},\ \{\texttt{crossTogether(joe, averell)}\}\ \rangle$$

## 3. Actions with Costs

Using the language $\mathcal{K}$ and the system prototype, $\texttt{DLV}^{\mathcal{K}}$, we can already express and solve some involved planning tasks, cf. (Eiter et al., 2003b). However, $\mathcal{K}$ and $\texttt{DLV}^{\mathcal{K}}$ alone offer no means for finding optimal plans under an objective cost function. In general, different criteria of plan optimality can be relevant, such as optimality wrt. action costs as shown in the next example, which is a slight elaboration of the Bridge Crossing Problem, and a well-known brain teasing riddle:

**Problem 2 [Quick Bridge Crossing Problem]** *The persons in the bridge crossing scenario need different times to cross the bridge, namely 1, 2, 5, and 10 minutes, respectively. Walking in two implies moving at the slower rate of both. Is it possible that all four persons get across within 17 minutes?*

On first thought this is infeasible, since the seemingly optimal plan where `joe`, who is the fastest, keeps the lamp and leads all the others across takes 19 minutes altogether. Surprisingly, as we will see, the optimal solution indeed only takes 17 minutes.

In order to allow for an elegant and convenient encoding of such optimization problems, we extend $\mathcal{K}$ to the language $\mathcal{K}^c$ in which one can assign costs to actions.

### 3.1 Syntax of $\mathcal{K}^c$

Let $\sigma^{act}$, $\sigma^{fl}$, and $\sigma^{var}$ denote (finite) sets of action names, fluent names and variable symbols. Furthermore, let $\mathcal{L}_{act}$, $\mathcal{L}_{fl}$, and $\mathcal{L}_{typ}$ denote the sets of action, fluent, and type literals, respectively,

2. This is also the format of the input files of our system prototype, which will be presented in Section 7.

```
actions :    cross(X) requires person(X).
             crossTogether(X,Y) requires person(X), person(Y), X < Y.
             takeLamp(X) requires person(X).

fluents :    across(X) requires person(X).
             differentSides(X,Y) requires person(X), person(Y).
             hasLamp(X) requires person(X).

initially :  -across(X). hasLamp(joe).

always :     executable crossTogether(X,Y) if hasLamp(X).
             executable crossTogether(X,Y) if hasLamp(Y).
             nonexecutable crossTogether(X,Y) if differentSides(X,Y).

             executable cross(X) if hasLamp(X).

             executable takeLamp(X).
             nonexecutable takeLamp(X) if hasLamp(Y), differentSides(X,Y).

             caused across(X) after crossTogether(X,Y), -across(X).
             caused across(Y) after crossTogether(X,Y), -across(Y).
             caused -across(X) after crossTogether(X,Y), across(X).
             caused -across(Y) after crossTogether(X,Y), across(Y).

             caused across(X) after cross(X), -across(X).
             caused -across(X) after cross(X), across(X).

             caused hasLamp(X) after takeLamp(X).
             caused -hasLamp(X) after takeLamp(Y), X != Y, hasLamp(X).

             caused differentSides(X,Y) if across(X), -across(Y).
             caused differentSides(X,Y) if -across(X), across(Y).

             inertial across(X).
             inertial -across(X).
             inertial hasLamp(X).

noConcurrency.

goal :       across(joe), across(jack), across(william), across(averell)? (l)
```

Figure 1: $\mathcal{K}$ encoding of the Bridge Crossing Problem

formed from the action names, fluent names, and predicates in the background knowledge (including built-in predicates), respectively, using terms from a nonempty (finite) set of constants $\sigma^{con}$.

$\mathcal{K}^c$ extends action declarations as in $\mathcal{K}$ with costs as follows.

**Definition 3.1** *An* action declaration $d$ *in* $\mathcal{K}^c$ *is of the form:*

$$p(X_1,\ldots,X_n) \texttt{ requires } t_1,\ldots,t_m \texttt{ costs } C \texttt{ where } c_1,\ldots,c_k. \tag{1}$$

*where (1)* $p \in \sigma^{act}$ *has arity* $n \geq 0$*, (2)* $X_1,\ldots,X_n \in \sigma^{var}$*, (3)* $t_1,\ldots,t_m$*,* $c_1,\ldots,c_k$ *are from* $\mathcal{L}_{typ}$ *such that every* $X_i$ *occurs in* $t_1,\ldots,t_m$*, (4)* $C$ *is either an integer constant, a variable from the set of all variables occurring in* $t_1,\ldots,t_m$*,* $c_1,\ldots,c_k$ *(denoted by* $\sigma^{var}(d)$*), or the distinguished variable* `time`*, (5)* $\sigma^{var}(d) \subseteq \sigma^{var} \cup \{\texttt{time}\}$*, and (6)* `time` *does not occur in* $t_1,\ldots t_m$*.*

If $m = 0$, the keyword '`requires`' is omitted; if $k = 0$, the keyword '`where`' is omitted and '`costs` $C$' is optional. Here, (1) and (2) state that parameters to an action must be variables, and not fixed values. Informally, (3) means that all parameters of an action must be "typed" in the `requires` part. Condition (4) asserts that the cost is locally defined or given by the stage of the plan, which is referenced through the global variable `time`. Conditions (5) and (6) ensure that all variables are known and that type information of action parameters is static, i.e., does not depend on time.

Planning domains and planning problems in $\mathcal{K}^c$ are defined as in $\mathcal{K}$.

For example, in the elaborated Bridge Crossing Problem, the declaration of $\mathtt{cross(X)}$ can be extended as follows: suppose a predicate $\mathtt{walk(Person, Minutes)}$ in the background knowledge indicates that `Person` takes `Minutes` to cross. Then, we may simply declare

$\mathtt{cross(X)}$ `requires` $\mathtt{person(X)}$ `costs` `WX` `where` $\mathtt{walk(X, WX)}$.

### 3.2 Semantics of $\mathcal{K}^c$

Semantically, $\mathcal{K}^c$ extends $\mathcal{K}$ by the cost values of actions at points in time. In any plan $P = \langle A_1, \ldots, A_l \rangle$, at step $1 \leq i \leq l$, the actions in $A_i$ are executed to reach time point $i$.

A ground action $p(x_1, \ldots, x_n)$ is a *legal action instance* of an action declaration $d$ wrt. a $\mathcal{K}^c$ planning domain $PD = \langle \Pi, \langle D, R \rangle \rangle$, if there exists some ground substitution $\theta$ for $\sigma^{var}(d) \cup \{\mathtt{time}\}$ such that $X_i\theta = x_i$, for $1 \leq i \leq n$ and $\{t_1\theta, \ldots, t_m\theta\} \subseteq M$, where $M$ is the unique answer set of the background knowledge $\Pi$. Any such $\theta$ is called a *witness substitution* for $p(x_1, \ldots, x_n)$. Informally, an action instance is legal, if it satisfies the respective typing requirements. Action costs are now formalized as follows.

**Definition 3.2** *Let $a = p(x_1, \ldots, x_n)$ be a legal action instance of a declaration $d$ of the form (1) and let $\theta$ be a witness substitution for $a$. Then*

$$cost_\theta(p(x_1, \ldots, x_n)) = \begin{cases} 0, & \textit{if the } \mathtt{costs} \textit{ part of } d \textit{ is empty;} \\ val(C\theta), & \textit{if } \{c_1\theta, \ldots, c_k\theta\} \subseteq M; \\ \textit{undefined} & \textit{otherwise.} \end{cases}$$

*where $M$ is the unique answer set of $\Pi$ and $val : \sigma^{con} \rightarrow \mathbb{N}$ is defined as the integer value for integer constants and 0 for all non-integer constants.*

By reference to the variable `time`, it is possible to define time-dependent action costs; we shall consider an example in Section 5.2. Using $cost_\theta$, we now introduce well-defined legal action instances and define action cost values as follows.

**Definition 3.3** *A legal action instance $a = p(x_1, \ldots, x_n)$ is* well-defined *iff it holds that (i) for any time point $i \geq 1$, there is some witness substitution $\theta$ for $a$ such that $\mathtt{time} = i$ and $cost_\theta(a)$ is an integer, and (ii) $cost_\theta(a) = cost_{\theta'}(a)$ holds for any two witness substitutions $\theta, \theta'$ which coincide on* `time` *and have defined costs. For any well-defined $a$, its unique cost at time point $i \geq 1$ is given by $cost_i(a) = cost_\theta(a)$ where $\theta$ is as in (i).*

In this definition, condition (i) ensures that some cost value exists, which must be an integer, and condition (ii) ensures that this value is unique, i.e., any two different witness substitutions $\theta$ and $\theta'$ for $a$ evaluate the `cost` part to the same integer cost value.

An action declaration $d$ is *well-defined*, if all its legal instances are well-defined. This will be fulfilled if, in database terms, the variables $X_1, \ldots, X_n$ together with `time` in (1) functionally determine the value of $C$. In our framework, the semantics of a $\mathcal{K}^c$ planning domain $PD = \langle \Pi, \langle D, R \rangle \rangle$ is only well-defined for well-defined action declarations in $PD$. In the rest of this paper, we assume well-definedness of $\mathcal{K}^c$ unless stated otherwise.

Using $cost_i$, we now define costs of plans.

**Definition 3.4** *Let $\mathcal{P} = \langle PD, Q\,?\,(l) \rangle$ be a planning problem. Then, for any plan $P = \langle A_1, \ldots, A_l \rangle$ for $\mathcal{P}$, its cost is defined as*

$$cost_{\mathcal{P}}(P) = \textstyle\sum_{j=1}^{l} \left( \sum_{a \in A_j} cost_j(a) \right).$$

*A plan $P$ is* optimal *for $\mathcal{P}$, if $cost_{\mathcal{P}}(P) \leq cost_{\mathcal{P}}(P')$ for each plan $P'$ for $\mathcal{P}$, i.e., $P$ has least cost among all plans for $\mathcal{P}$. The cost of a planning problem $\mathcal{P}$, denoted $cost^*_{\mathcal{P}}$, is given by $cost^*_{\mathcal{P}} = cost_{\mathcal{P}}(P^*)$, where $P^*$ is an optimal plan for $\mathcal{P}$.*

In particular, $cost_{\mathcal{P}}(P) = 0$ if $P = \langle \rangle$, i.e., the plan is void. Note that $cost^*_{\mathcal{P}}$ is only defined if a plan for $\mathcal{P}$ exists.[3]

Usually one only can estimate some *upper bound* of the plan length, but does not know the exact length of an optimal plan. Although we have only defined optimality for a fixed plan length $l$, we will see in Section 5.1 that by appropriate encodings this can be extended to optimality for plans with length *at most* $l$.

Besides optimal plans, also plans with bounded costs are of interest, which motivates the following definition.

**Definition 3.5** *A plan $P$ for a planning problem $\mathcal{P}$ is* admissible *wrt. cost $c$, if $cost_{\mathcal{P}}(P) \leq c$.*

Admissible plans impose a weaker condition on the plan quality than optimal plans. They are particularly relevant if optimal costs are not a crucial issue, as long as the cost stays within a given limit, and if optimal plans are difficult to compute. We might face questions like “Can I make it to the airport within one hour?”, “Do I have enough change to buy a coffee?” etc. which amount to admissible planning problems. As we shall see, computing admissible plans is complexity-wise easier than computing optimal plans.

### 3.3 An Optimal Solution for the Quick Bridge Crossing Problem

To model the Quick Bridge Crossing Problem in $\mathcal{K}^c$, we first extend the background knowledge as follows, where the predicate ‘`walk`’ describes the time a person needs to cross and ‘`max`’ determines which of two persons is slower:

```
walk(joe, 1). walk(jack, 2). walk(william, 5). walk(averell, 10).
max(A, B, A) :- walk(_, A), walk(_, B), A >= B.
max(A, B, B) :- walk(_, A), walk(_, B), B > A.
```

Next, we modify the declarations for `cross` and `crossTogether` from Figure 1 by adding costs:

3. In the following, subscripts will be dropped when clear from the context.

```
cross(X) requires person(X) costs WX where walk(X,WX).
crossTogether(X,Y) requires person(X), person(Y), X < Y
                   costs Wmax where walk(X,WX), walk(Y,WY), max(WX,WY,Wmax).
```

The declaration of `takeLamp` remains unchanged, as the time to hand over the lamp is negligible.

Using this modified planning domain, the 5-step plan reported in Section 2.1 has cost 19. Actually, it is optimal for plan length $l = 5$. However, when we relinquish the first intuition that the fastest person, `joe`, always has the lamp and consider the problem under varying plan length, then we can find the following 7-step plan:

$$P = \langle\ \{\texttt{crossTogether(joe,jack)}\},\ \{\texttt{cross(joe)}\},\ \{\texttt{takeLamp(william)}\},\ \{\texttt{crossTogether(william,averell)}\},\ \{\texttt{takeLamp(jack)}\},\ \{\texttt{cross(jack)}\},\ \{\texttt{crossTogether(joe,jack)}\}\ \rangle$$

Here, $cost_{\mathcal{P}}(P) = 17$, and thus $P$ is admissible with respect to cost 17. This means that the Quick Bridge Crossing Problem has a positive answer. In fact, $P$ has least cost over all plans of length $l = 7$, and is thus an optimal 7-step plan. Moreover, $P$ has also least cost over all plans that emerge if we consider all plan lengths. Thus, $P$ is an optimal solution for the Quick Bridge Crossing Problem under arbitrary plan length.

### 3.4 Bridge Crossing under Incomplete Knowledge

The language $\mathcal{K}$ is well-suited to model problems which involve uncertainty such as incomplete initial states or non-deterministic action effects at a qualitative level. The enriched language $\mathcal{K}^c$ gracefully extends to secure (conformant) plans as well, which must reach the goal under all circumstances (Eiter et al., 2000b, 2003b). More precisely, an optimistic plan $\langle A_1, \ldots, A_n\rangle$ is *secure*, if it is applicable under any evolution of the system: starting from any legal initial state $s_0$, the first action set $A_1$ (for plan length $l \geq 1$) can always be executed (i.e., some legal transition $\langle s_0, A_1, s_1\rangle$ exists), and for every such possible state $s_1$, the next action set $A_2$ can be executed etc., and after having performed all actions, the goal is always accomplished (cf. Appendix for a formal definition).

While secure plans inherit costs from optimistic plans, there are different possibilities to define optimality of secure plans. We may consider a secure plan as optimal, if it has least cost either

- among all optimistic plans, or
- among all secure plans only.

In the first alternative, there might be planning problems which have secure plans, but no optimal secure plans. For this reason, the second alternative appears to be more appropriate.

**Definition 3.6** *A secure plan $P$ is* optimal *for a planning problem $\mathcal{P}$, if it has least cost among all secure plans for $\mathcal{P}$, i.e., $cost_{\mathcal{P}}(P) \leq cost_{\mathcal{P}}(P')$ for each secure plan $P'$ for $\mathcal{P}$. The* secure cost *of $\mathcal{P}$, denoted $cost^*_{sec}(\mathcal{P})$, is $cost^*_{sec}(\mathcal{P}) = cost_{\mathcal{P}}(P^*)$, where $P^*$ is any optimal secure plan for $\mathcal{P}$.*

The notion of admissible secure plans is defined analogously.

For example, assume that it is known that at least one person in the bridge scenario has a lamp, but that neither the exact number of lamps nor the allocation of lamps to persons is known. If the four desperate persons now ask for a plan which brings them safely across the bridge, we need a (fast) secure plan that works under all possible initial situations. In $\mathcal{K}^c$, this can be modeled by replacing the `initially`-part with the following declarations:

```
initially: total hasLamp(X).
           caused false if -hasLamp(joe), -hasLamp(jack),
                           -hasLamp(william), -hasLamp(averell).
```

The first statement says that each person either has a lamp or not, and the second that at least one of them must have a lamp. For a detailed discussion on the use of the "total" statement for modeling incomplete knowledge and non-determinism we refer to (Eiter et al., 2003b).

As we can easily see, an optimal secure solution will take at least 17 minutes, since the original case (where only joe has a lamp) is one of the possible initial situations, for which the cost of an optimistic plan which is optimal over all plan lengths was 17. However, a secure plan which is optimal over all plan lengths requires at least 8 steps now (but no higher cost): Different from optimistic plans, we need one extra step at the beginning which makes sure that one of those who walk first (above, joe and jack) has the lamp, which is effected by the proper takeLamp action.

An example of such a plan is the following which has cost 17:

$$P = \langle\ \{\texttt{takeLamp(joe)}\},\ \{\texttt{crossTogether(joe,jack)}\},\ \{\texttt{cross(joe)}\},\ \{\texttt{takeLamp(william)}\},\ \{\texttt{crossTogether(william,averell)}\}, \{\texttt{takeLamp(jack)}\},\ \{\texttt{cross(jack)}\},\ \{\texttt{crossTogether(joe,jack)}\}\ \rangle$$

We can easily check that $P$ works for every possible initial situation. Thus, it is an optimal (secure) plan for plan length 8, and moreover also for arbitrary plan length.

## 4. Computational Complexity

In this section, we will address the computational complexity of $\mathcal{K}^c$, complementing similar results for the language $\mathcal{K}$ (Eiter et al., 2003b).

### 4.1 Complexity Classes

We assume that the reader is familiar with the basic notions of complexity theory, such as P, NP, problem reductions and completeness; see e.g. (Papadimitriou, 1994) and references therein. We recall that the Polynomial Hierarchy (PH) contains the classes $\Sigma_0^P = \Pi_0^P = \Delta_0^P = \mathrm{P}$ and $\Sigma_{i+1}^P = \mathrm{NP}^{\Sigma_i^P}$, $\Pi_{i+1}^P = \text{co-}\Sigma_{i+1}^P$, $\Delta_{i+1}^P = \mathrm{P}^{\Sigma_i^P}$, for $i \geq 0$. In particular, $\Sigma_1^P = \mathrm{NP}$ and $\Delta_2^P = \mathrm{P}^{\mathrm{NP}}$. Note that these classes contain decision problems (i.e., problems where the answer is "yes" or "no"). While checking well-definedness and deciding plan existence are such problems, computing a plan is a *search problem*, where for each problem instance $I$ a (possibly empty) finite set $S(I)$ of solutions exists. To solve such a problem, a (possibly nondeterministic) algorithm must compute the alternative solutions from this set in its computation branches, if $S(I)$ is not empty. More precisely, search problems are solved by transducers, i.e., Turing machines equipped with an output tape. If the machine halts in an accepting state, then the contents of the output tape is the result of the computation. Observe that a nondeterministic machine computes a (partial) multi-valued function.

As an analog to NP, the class NPMV contains those search problems where $S(I)$ can be computed by a nondeterministic Turing machine in polynomial time; for a precise definition, see (Selman, 1994). In analogy to $\Sigma_{i+1}^P$, by $\Sigma_{i+1}^P\mathrm{MV} = \mathrm{NPMV}^{\Sigma_i^P}$, $i \geq 0$, we denote the generalization of NPMV where the machine has access to a $\Sigma_i^P$ oracle.

Analogs to the classes P and $\Delta_{i+1}^P$, $i \geq 0$, are given by the classes FP and $\mathrm{F}\Delta_{i+1}^P$, $i \geq 0$, which contain the partial single-valued functions (that is, $|S(I)| \leq 1$ for each problem instance $I$) computable in polynomial time using no resp. a $\Sigma_i^P$ oracle. We say, abusing terminology, that a search problem $A$ is in FP (resp. $\mathrm{F}\Delta_{i+1}^P$), if there is a partial (single-valued) function $f \in \mathrm{FP}$ (resp. $f \in \mathrm{F}\Delta_{i+1}^P$) such that $f(I) \in S(I)$ and $f(I)$ is undefined iff $S(I) = \emptyset$. For example, computing a satisfying assignment for a propositional CNF (FSAT) and computing an optimal tour in the Traveling Salesperson Problem (TSP) are in $\mathrm{F}\Delta_2^P$ under this view, cf. (Papadimitriou, 1994).

A partial function $f$ is polynomial-time reducible to another partial function $g$, if there are polynomial-time computable functions $h_1$ and $h_2$ such that $f(I) = h_2(I, g(h_1(I)))$ for all $I$ and $g(h_1(I))$ is defined whenever $f(I)$ is defined. Hardness and completeness are defined as usual.

### 4.2 Problem Setting

We will focus on the following questions:

**Checking Well-Definedness:** Decide whether a given action description is well-defined wrt. a given planning domain $PD$, resp. whether a given planning domain $PD$ is well-defined.

**Admissible Planning:** Decide whether for planning problem $\mathcal{P}$ an admissible (optimistic/secure) plan exists wrt. a given cost value $c$, and find such a plan.

**Optimal Planning:** Find an optimal (optimistic/secure) plan for a given planning problem.

Notice that (Eiter et al., 2003b) focused on deciding the existence of optimistic/secure plans, rather than on actually finding plans, and presented a detailed study of the complexity of this task under various restrictions for ground (propositional) planning problems. In this paper, we confine the discussion to the case of planning problems $\mathcal{P} = \langle PD, Q\,?\,(l)\rangle$ which look for *polynomial length plans*, i.e., problems where the plan length $l$ is bounded by some polynomial in the size of the input.

We shall consider here mainly ground (propositional) planning, and assume that the planning domains are well-typed and that the unique model of the background knowledge can be computed in polynomial time. In the general case, by well-known complexity results on logic programming, cf. (Dantsin, Eiter, Gottlob, & Voronkov, 2001), already evaluating the background knowledge is EXPTIME-hard, and the problems are thus provably intractable. We recall the following results, which appear in (or directly follow from) previous work (Eiter et al., 2003b).

**Proposition 4.1** *Deciding, given a propositional planning problem $\mathcal{P}$ and a sequence $P = \langle A_1, \ldots, A_l\rangle$ of action sets, (i) whether a given sequence $T = \langle t_1, \ldots, t_l\rangle$ is a legal trajectory witnessing that $P$ is an optimistic plan for $\mathcal{P}$ is feasible in polynomial time, and (ii) whether $P$ is a secure plan for $\mathcal{P}$ is $\Pi_2^P$-complete.*

### 4.3 Results

We start by considering checking well-definedness. For this problem, it is interesting to investigate the non-ground case, assuming that the background knowledge is already evaluated. This way we can assess the intrinsic difficulty of this task obtaining the following result.

**Theorem 4.2 (Complexity of checking well-definedness)** *Given a $\mathcal{K}^c$ planning domain $PD = \langle \Pi, \langle D, R\rangle\rangle$ and the unique model $M$ of $\Pi$, checking (i) well-definedness of a given action declaration $d$ of form (1) wrt. $PD$ and (ii) well-definedness of $PD$ are both $\Pi_2^P$-complete.*

*Proof.* *Membership:* As for (i), $d$ is violated if it has a nonempty `costs` part and a legal action instance $a = p(x_1, \ldots, x_n)$ such that either (1) there exist witness substitutions $\theta$ and $\theta'$ for $a$ such that $\mathtt{time}\theta = \mathtt{time}\theta'$, $cost_\theta(a) = val(C\theta)$ and $cost_{\theta'}(a) = val(C\theta')$, and $val(C\theta) \neq val(C\theta')$, or (2) there is no witness substitution $\theta$ for $a$ such that $cost_\theta(a) = val(C\theta)$ is an integer. Such an $a$ can be guessed and checked, via a witness substitution, in polynomial time, and along with $a$ also $\theta$ and $\theta'$ as in (1); note that, by definition, all variables must be substituted by constants from the background knowledge (including numbers), and so must be values for `time` if it occurs in $c_1, \ldots, c_k$. Given $a$, we can decide (2) with the help of an NP oracle. In summary, disproving well-definedness of $d$ is nondeterministically possible in polynomial time with an NP oracle. Hence, checking well-definedness of $d$ is in co-$\Sigma_2^P$ = $\Pi_2^P$. The membership part of (ii) follows from (i), since well-definedness of *PD* reduces to well-definedness of all action declarations in it, and $\Pi_2^P$ is closed under conjunctions.
*Hardness:* We show hardness for (i) by a reduction from deciding whether a quantified Boolean formula (QBF)

$$Q = \forall X \exists Y. c_1 \wedge \cdots \wedge c_k$$

where each $c_i = L_{i,1} \vee \cdots \vee L_{i,\ell_i}$, $i = 1, \ldots, k$, is a disjunction of literals $L_{i,j}$ on the atoms $X = x_1, \ldots, x_n$ and $Y = x_{n+1} \ldots, x_m$, is true. Without loss of generality, we may assume that each $c_i$ contains three (not necessarily distinct) literals, which are either all positive or all negative.

We construct a planning domain *PD* and $d$ as follows. The background knowledge, $\Pi$, is

```
bool(0). bool(1).
pos(1,0,0). pos(0,1,0). pos(0,0,1). pos(1,1,0). pos(1,0,1). pos(0,1,1). pos(1,1,1).
neg(0,0,0). neg(1,0,0). neg(0,1,0). neg(0,0,1). neg(1,1,0). neg(1,0,1). neg(0,1,1).
```

Here, `bool` declares the truth values 0 and 1. The facts $\mathtt{pos}(\mathtt{X_1}, \mathtt{X_2}, \mathtt{X_3})$ and $\mathtt{neg}(\mathtt{X_1}, \mathtt{X_2}, \mathtt{X_3})$ state those truth assignments to $X_1$, $X_2$, and $X_3$ such that the positive clause $X_1 \vee X_2 \vee X_3$ resp. the negative clause $\neg X_1 \vee \neg X_2 \vee \neg X_3$ is satisfied.

The rest of the planning domain *PD* consists of the single action declaration $d$ of form

$\mathtt{p(V_1, ..., V_n)}$ `requires` $\mathtt{bool(V_1), ..., bool(V_n)}$ `costs 0 where` $c_1^*, ..., c_k^*$.

where

$$c_i^* = \begin{cases} \mathtt{pos}(\mathtt{V_{i,1}}, \mathtt{V_{i,2}}, \mathtt{V_{i,3}}), & \text{if } c_i = x_{i,1} \vee x_{i,2} \vee x_{i,3}, \\ \mathtt{neg}(\mathtt{V_{i,1}}, \mathtt{V_{i,2}}, \mathtt{V_{i,3}}), & \text{if } c_i = \neg x_{i,1} \vee \neg x_{i,2} \vee \neg x_{i,3}, \end{cases} \qquad i = 1, \ldots, k.$$

For example, the clause $c = x_1 \vee x_3 \vee x_6$ is mapped to $c^* = \mathtt{pos}(\mathtt{V_1}, \mathtt{V_3}, \mathtt{V_6})$. It is easy to see that each legal action instance $a = p(b_1, \ldots, b_n)$ of $d$ corresponds 1-1 to the truth assignment $\sigma_a$ of $X$ given by $\sigma_a(x_i) = b_i$, for $i = 1, \ldots, n$. Furthermore, $a$ has a cost value defined (which is 0) iff the formula $\exists Y (c_1\sigma_a \wedge \cdots \wedge c_k\sigma_a)$ is true. Thus, $d$ is well-defined wrt. *PD* iff $Q$ is true. Since *PD* and $d$ are efficiently constructible, this proves $\Pi_2^P$-hardness. □

Observe that in the ground case, checking well-definedness is much easier. Since no substitutions need to be guessed, the test in the proof of Theorem 4.2 is polynomial. Thus, by our assumption on the efficient evaluation of the background program, we obtain:

**Corollary 4.3** *In the ground (propositional) case, checking well-definedness of an action description $d$ wrt. a $\mathcal{K}^c$ planning domain PD = $\langle \Pi, \langle D, R \rangle \rangle$, resp. of PD as a whole, is possible in polynomial time.*

We remark that checking well-definedness can be expressed as a planning task in $\mathcal{K}$, and also by a logic program; we refer to (Eiter, Faber, Leone, Pfeifer, & Polleres, 2002b) for details.

We now turn to computing admissible plans.

**Theorem 4.4 (Complexity of admissible planning)** *For polynomial plan lengths, deciding whether a given (well-defined) propositional planning problem $\langle PD, q \rangle$ has (i) some optimistic admissible plan wrt. to a given integer $b$ is $\mathrm{NP}$-complete, and finding such a plan is complete for $\mathrm{NPMV}$, (ii) deciding whether $\langle PD, q \rangle$ has some secure admissible plan wrt. to a given integer $b$ is $\Sigma_3^P$-complete, and computing such a plan is $\Sigma_3^P\mathrm{MV}$-complete. Hardness holds in both cases for fixed plan length.*

As for the proof we refer to the Appendix. We finally address the complexity of computing optimal plans.

**Theorem 4.5 (Complexity of optimal planning)** *For polynomial plan lengths, (i) computing an optimal optimistic plan for $\langle PD, Q\,?\,(l) \rangle$ in $\mathcal{K}^c$ is $\mathrm{F}\Delta_2^P$-complete, and (ii) computing an optimal secure plan for $\langle PD, Q\,?\,(l) \rangle$ in $\mathcal{K}^c$ is $\mathrm{F}\Delta_4^P$-complete. Hardness holds in both cases even if the plan length $l$ is fixed.*

The proof again can be found in the in the Appendix.

We remark that in the case of unbounded plan length, the complexity of computing plans increases and requires (at least) exponential time in general, since plans might have exponential length in the size of the planning problem. Thus, in practical terms, constructing such plans is infeasible, since they occupy exponential space. Furthermore, as follows from previous results (Eiter et al., 2003b), deciding the existence of an admissible optimistic resp. secure plan for a planning problem wrt. a given cost is PSPACE-complete resp. NEXPTIME-complete. We leave a more detailed analysis of complexity aspects of $\mathcal{K}^c$ for further work.

## 5. Applications

### 5.1 Cost Efficient versus Time Efficient Plans

In this section, we show how the language $\mathcal{K}^c$ can be used to minimize plan length in combination with minimizing the costs of a plan. This is especially interesting in problem settings where parallel actions are allowed (cf. (Kautz & Walser, 1999; Lee & Lifschitz, 2001)).

For such domains with parallel actions, Kautz and Walser propose various criteria to be optimized, for instance the number of actions needed, or the number of necessary time steps when parallel actions are allowed, as well as combinations of these two criteria (1999). By exploiting action costs and proper modeling, we can solve optimization problems of this sort. For example, we can single out plans with a minimal number of actions simply by assigning cost 1 to all possible actions.

We consider the following optimization problems:

($\alpha$) Find a plan with minimal cost (*cheapest plan*) for a given number of steps.

($\beta$) Find a plan with minimal time steps (*shortest plan*).

($\gamma$) Find a shortest among the cheapest plans.

($\delta$) Find a cheapest among the shortest plans.

Problem ($\alpha$) is what we have already defined as optimal plans so far. We will now show how to express ($\beta$) in terms of optimal cost plans as well, and how to extend this elaboration with respect to the combinations ($\gamma$) and ($\delta$).

#### 5.1.1 CHEAPEST PLANS WITH GIVEN PLAN LENGTH ($\alpha$)

As a guiding example, we refer to Blocks World with parallel moves allowed, where apart from finding shortest plans also minimizing the total number of moves is an issue. A $\mathcal{K}^c$ encoding for this domain, where plans are serializable, is shown in Figure 2. Serializability here means that parallel actions are non-interfering and can be executed sequentially in any order, i.e. the parallel plan can be arbitrarily "unfolded" to a sequential plan.

```
fluents:  on(B,L) requires block(B), location(L).
          blocked(B) requires block(B).
          moved(B) requires block(B).

actions:  move(B,L) requires block(B), location(L) costs 1.

always:   executable move(B,L) if B != L.
          nonexecutable move(B,L) if blocked(B).
          nonexecutable move(B,L) if blocked(L).
          nonexecutable move(B,L) if move(B1,L), B < B1, block(L).
          nonexecutable move(B,L) if move(B,L1), L < L1.
          nonexecutable move(B,B1) if move(B1,L).

          caused on(B,L) after move(B,L).
          caused blocked(B) if on(B1,B).
          caused moved(B) after move(B,L).
          caused on(B,L) if not moved(B) after on(B,L).
```

Figure 2: $\mathcal{K}^c$ encoding for the Blocks World domain

The planning problem emerging from the initial state and the goal state depicted in Figure 3 can be modeled using the background knowledge $\Pi_{bw}$:

```
block(1). block(2). block(3). block(4). block(5). block(6).
location(table).
location(B) :- block(B).
```

and extending the program in Figure 2 as follows:

```
initially: on(1,2). on(2,table). on(3,4). on(4,table). on(5,6). on(6,table).
goal:      on(1,3), on(3,table), on(2,4), on(4,table), on(6,5), on(5,table) ?(l)
```

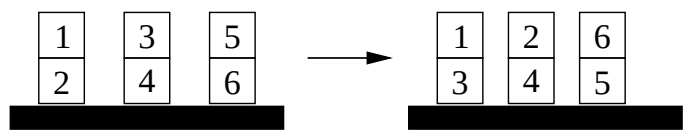

Figure 3: A simple Blocks World instance

Each move is penalized with cost 1, which results in a minimization of the total number of moves. Let $\mathcal{P}_l$ denote the planning problem for plan length $l$.

For $l = 2$, we have an optimal plan which involves six moves, i.e. $cost^*_{\mathcal{P}_2} = 6$:

$$P_2 = \langle\, \{\mathtt{move(1, table)}, \mathtt{move(3, table)}, \mathtt{move(5, table)}\},\ \ \{\mathtt{move(1, 3)}, \mathtt{move(2, 4)}, \mathtt{move(6, 5)}\}\,\rangle$$

By unfolding the steps, this plan gives rise to similar plans of length $l = 3, \ldots, 6$ that have cost 6. For $l = 3$, we can find among others the following optimal plan, which has cost 5:

$$P_3 = \langle\, \{\mathtt{move(3, table)}\},\ \ \{\mathtt{move(1, 3)}, \mathtt{move(5, table)}\},\ \ \{\mathtt{move(2, 4)}, \mathtt{move(6, 5)}\}\,\rangle$$

This plan can not be further parallelized to having only two steps. For any plan length $l > 3$, we will obtain optimal plans similar to $P_3$, extended by void steps. Thus a plan which is cheapest over all plan lengths has cost 5 and needs three steps. Note that shortest parallel plans (of length 2) are more expensive, as explained above.

#### 5.1.2 SHORTEST PLANS ($\beta$)

Intuitively, it should be possible to include the minimization of time steps in the cost function. We describe a preprocessing method which, given a $\mathcal{K}$ planning domain $PD$, a list $Q$ of ground literals, and an upper bound $i \geq 0$ for the plan length, generates a planning problem $\mathcal{P}_\beta(PD, Q, i)$ such that the optimal plans for $\mathcal{P}_\beta$ correspond to shortest plans which reach $Q$ in $PD$ in at most $i$ steps, i.e., to plans for $\langle PD, Q\,?\,(l)\rangle$ such that $l \leq i$ is minimal. We assume that no action costs are specified in the original planning domain $PD$, and minimizing time steps is our only target.

First we rewrite the planning domain $PD$ to $PD_\beta$ as follows: We introduce a new distinct fluent `gr` and a new distinct action `finish`, defined as follows:

```
fluents :   gr.
actions :   finish costs time.
```

Intuitively, the action `finish` represents a final action, which we use to finish the plan. The later this action occurs, the more expensive the plan as we assign `time` as cost. The fluent `gr` ("goal reached") shall be true and remain true as soon as the goal has been reached, and it is triggered by the `finish` action.

This can be modeled in $\mathcal{K}^c$ by adding the following statements to the `always` section of the program:

```
executable finish if Q, not gr.
caused gr after finish.
caused gr after gr.
```

Furthermore, we want `finish` to occur exclusively and we want to block the occurrence of any other action once the goal has been reached. Therefore, for every action `A` in $PD$, we add

```
nonexecutable A if finish.
```

and add `not gr` to the `if`-part of each executability condition for `A`. Finally, to avoid any inconsistencies from static or dynamic effects as soon as the goal has been reached, we add `not gr` to the `if` part of any causation rule of the $PD$ except `nonexecutable` rules which remain unchanged.[4]

We define now $\mathcal{P}_\beta(PD, Q, i) = \langle PD_\beta, \mathtt{gr}\,?(i+1)\rangle$. We take $i + 1$ as the plan length since we need one additional step to execute the `finish` action.

4. There is no need to rewrite `nonexecutable` rules because the respective actions are already "switched off" by rewriting the executability conditions.

By construction, it is easy to see that any optimal plan $P = \langle A_1, \dots, A_j, A_{j+1}, \dots, A_{i+1}\rangle$ for the planning problem $\mathcal{P}_\beta$ must have $A_{j+1} = \{\texttt{finish}\}$ and $A_{j+2} = \dots = A_{i+1} = \emptyset$ for some $j \in \{0, \dots, i\}$. We thus have the following desired property.

**Proposition 5.1** *The optimal plans for $\mathcal{P}_\beta$ are in 1-1 correspondence to the shortest plans reaching $Q$ in $PD$. More precisely, $P = \langle A_1, \dots, A_{j+1}, \emptyset, \dots, \emptyset\rangle$ is an optimal optimistic plan for $\mathcal{P}_\beta(PD, Q, i)$ and $A_{j+1} = \{\texttt{finish}\}$ if and only if $P' = \langle A_1, \dots, A_j\rangle$ is an optimistic plan for $\langle PD, Q\,?\,(j)\rangle$ where $j \in \{0, \dots, i\}$, and $\langle PD, Q\,?\,(j')\rangle$ has no optimistic plan for each $j' < j$.*

In our Blocks World example, using this method we get all 2-step plans, if we choose $i \geq 2$.

To compute shortest plans over all plan lengths, we can set the upper bound $i$ large enough such that plans of length $l \leq i$ are guaranteed to exist. A trivial such bound is the total number of legal states which is in general exponential in the number of fluents.

However, many typical applications have an inherent, much smaller bound on the plan length. For instance, in a Blocks World with $n$ blocks, any goal configuration can be reached within at most $2n - s_{init} - s_{goal}$ steps, where $s_{init}$ and $s_{goal}$ are the numbers of stacks in the initial and the goal state, respectively.[5] Therefore, 6 is an upper bound for the plan length of our simple instance.

We remark that this approach for minimizing plan length is only efficient if an upper bound close to the optimum is known. Searching for a minimum length plan by iteratively increasing the plan length may be much more efficient if no such bound is known, since a weak upper bound can lead to an explosion of the search space (cf. the benchmarks in Section 7.2).

#### 5.1.3 SHORTEST AMONG THE CHEAPEST PLANS ($\gamma$)

In the previous subsection, we have shown how to calculate shortest plans for $\mathcal{K}$ programs without action costs. Combining arbitrary $\mathcal{K}^c$ programs and the rewriting method described there is easy. If we want to find a shortest among the cheapest plans, we can use the same rewriting, with just a little change. All we have to do is setting the costs of all actions except `finish` at least as high as the highest possible cost of the `finish` action. This is is obviously the plan length $i + 1$. So, we simply modify all action declarations

```
A requires B costs C where D.
```

in $\mathcal{P}_\beta$ by multiplying the costs with factor $i + 1$:

$$\texttt{A requires B costs C}_1\texttt{ where C}_1 = (i+1) * \texttt{C, D.}$$

This lets all other action costs take priority over the cost of `finish` and we can compute plans satisfying criterion ($\gamma$). Let $\mathcal{P}_\gamma$ denote the resultant planning problem. Then we have:

**Proposition 5.2** *The optimal plans for $\mathcal{P}_\gamma$ are in 1-1 correspondence to the shortest among the cheapest plans reaching $Q$ in $PD$ within $i$ steps. More precisely, $P = \langle A_1, \dots, A_{j+1}, \emptyset, \dots, \emptyset\rangle$ is an optimal optimistic plan for $\mathcal{P}_\gamma(PD, Q, i)$ and $A_{j+1} = \{\texttt{finish}\}$ if and only if (i) $P' = \langle A_1, \dots, A_j\rangle$ is a plan for $\mathcal{P}_j = \langle PD, Q\,?\,(j)\rangle$, where $j \in \{0, \dots, i\}$, and (ii) if $P'' = \langle A_1, \dots, A_{j'}\rangle$ is any plan for $\mathcal{P}_{j'} = \langle PD, Q\,?\,(j')\rangle$ where $j' \leq i$, then either $cost_{\mathcal{P}_{j'}}(P'') > cost_{\mathcal{P}_j}(P')$ or $cost_{\mathcal{P}_{j'}}(P'') = cost_{\mathcal{P}_j}(P')$ and $j' \geq j$.*

Figure 4 shows $\mathcal{P}_\gamma$ for our Blocks World instance where $i = 6$. One optimal plan for $\mathcal{P}_\gamma$ is

5. One can solve any Blocks World problem sequentially by first unstacking all blocks which are not on the table ($n - s_{init}$ steps) and then building up the goal configuration ($n - s_{goal}$ steps).

```
fluents :    on(B,L) requires block(B), location(L).
             blocked(B) requires block(B).
             moved(B) requires block(B).
             gr.

actions :    move(B,L) requires block(B), location(L) costs C where C = 7 * 1.
             finish costs time.

always :     executable move(B,L) if B != L, not gr.
             nonexecutable move(B,L) if blocked(B).
             nonexecutable move(B,L) if blocked(L).
             nonexecutable move(B,L) if move(B1,L), B < B1, block(L).
             nonexecutable move(B,L) if move(B,L1), L < L1.
             nonexecutable move(B,B1) if move(B1,L).

             caused on(B,L) if not gr after move(B,L).
             caused blocked(B) if on(B1,B), not gr.
             caused moved(B) if not gr after move(B,L).
             caused on(B,L) if not moved(B), not gr after on(B,L).

             executable finish if on(1,3), on(3,table), on(2,4), on(4,table),
                                  on(6,5), on(5,table), not gr.
             caused gr after finish.
             caused gr after gr.
             nonexecutable move(B,L) if finish.

initially :  on(1,2). on(2,table). on(3,4). on(4,table). on(5,6). on(6,table).

goal :       gr? (7)
```

Figure 4: Computing the shortest plan for a Blocks World instance with a minimum number of actions

$P = \langle$ $\{$move(3,table)$\}$, $\{$move(1,3),move(5,table)$\}$,
$\{$move(2,4),move(6,5)$\}$, $\{$finish$\}$, $\emptyset$, $\emptyset$, $\emptyset$ $\rangle$,

which has $cost_{\mathcal{P}_\gamma}(P) = 39$. We can now compute the optimal cost wrt. optimization ($\gamma$) by subtracting the cost of `finish` and dividing by $i+1$: $(39-4) \div (i+1) = 35 \div 7 = 5$. Thus, we need a minimum of 5 moves to reach the goal. The minimal number of steps is obviously all steps, except the final `finish` action, i.e. 3. Thus, we need at least 3 steps for a plan with five moves.

#### 5.1.4 CHEAPEST AMONG THE SHORTEST PLANS ($\delta$)

Again, we can use the rewriting for optimization ($\beta$). The cost functions have to be adapted similarly as in the previous subsection, such that now the cost of the action `finish` takes priority over all other actions costs. To this end, it is sufficient to set the cost of `finish` high enough, which is achieved by multiplying it with a factor $F$ higher than the sum of all action costs of all legal action instances at all steps $j = 1, \ldots, i+1$. Let $\mathcal{P}_\delta$ denote the resulting planning problem. We have:

**Proposition 5.3** *The optimal plans for $\mathcal{P}_\delta$ are in 1-1 correspondence to the cheapest among the shortest plans reaching $Q$ in PD within $i$ steps. More precisely,* $P = \langle A_1, \ldots, A_{j+1}, \emptyset, \ldots, \emptyset\rangle$ *is an optimal optimistic plan for* $\mathcal{P}_\delta(PD, Q, i)$ *and* $A_{j+1} = \{\texttt{finish}\}$ *if and only if (i)* $P' = \langle A_1, \ldots, A_j \rangle$ *is a plan for* $\mathcal{P}_j = \langle PD, Q\,?\,(j) \rangle$*, where* $j \in \{0, \ldots, i\}$*, and (ii) if* $P'' = \langle A_1, \ldots, A_{j'} \rangle$ *is any plan for* $\mathcal{P}_{j'} = \langle PD, Q\,?\,(j') \rangle$ *where* $j' \leq i$*, then either* $j' > j$*, or* $j' = j$ *and* $cost_{\mathcal{P}_{j'}}(P'') \geq cost_{\mathcal{P}_j}(P')$.

In our example, there are 36 possible moves. Thus, we could take $F = 36 * (i + 1)$ and would set the costs of `finish` to $\texttt{time} * 36 * (i + 1)$. However, we only need to take into account those actions which can actually occur simultaneously. In our example, at most six blocks can be moved in parallel. Therefore, it is sufficient to set $F = 6 * (i + 1)$ and assign `finish` cost $\texttt{time} * F = \texttt{time} * 42$. Accordingly, the action declarations are modified as follows:

```
actions :   move(B, L) requires block(B), location(L) costs 1.
            finish costs C where C = time * 42.
```

An optimal plan for the modified planning problem $\mathcal{P}_\delta$ is:

$$P = \langle\ \{\texttt{move}(1,\texttt{table}), \texttt{move}(3,\texttt{table}), \texttt{move}(5,\texttt{table})\},\ \{\texttt{move}(1,3), \texttt{move}(2,4), \texttt{move}(6,5)\},\ \{\texttt{finish}\},\ \emptyset,\ \emptyset,\ \emptyset,\ \emptyset \rangle$$

We have $cost_{\mathcal{P}_\delta}(P) = 132$. Here, we can compute the optimal cost wrt. optimization ($\delta$) by simply subtracting the cost of `finish`, i.e. $132 - 3 * 42 = 6$, since `finish` occurs at time point $3$. Consequently, we need a minimum of $6$ moves for a shortest plan, which has length $3 - 1 = 2$.

And indeed, we have seen that (and how) the optimization problems ($\alpha$) through ($\delta$) can be represented in $\mathcal{K}^c$. We remark that the transformations $\mathcal{P}_\beta$, $\mathcal{P}_\gamma$, and $\mathcal{P}_\delta$ all work under the restrictions to secure and/or sequential plans as well.

### 5.2 Traveling Salesperson

As another illustrating example for optimal cost planning, we will now introduce some elaboration of the Traveling Salesperson Problem.

**Traveling Salesperson Problem (TSP).** We start with the classical Traveling Salesperson Problem (TSP), where we have a given set of cities and connections (e.g., roads, airways) of certain costs. We want to know a most economical round trip which visits all cities exactly once and returns to the starting point (if such a tour exists). Figure 5 shows an instance representing the capitals of all Austrian provinces. The dashed line is a flight connection, while all other connections are roads; each connection is marked with the costs in traveling hours.

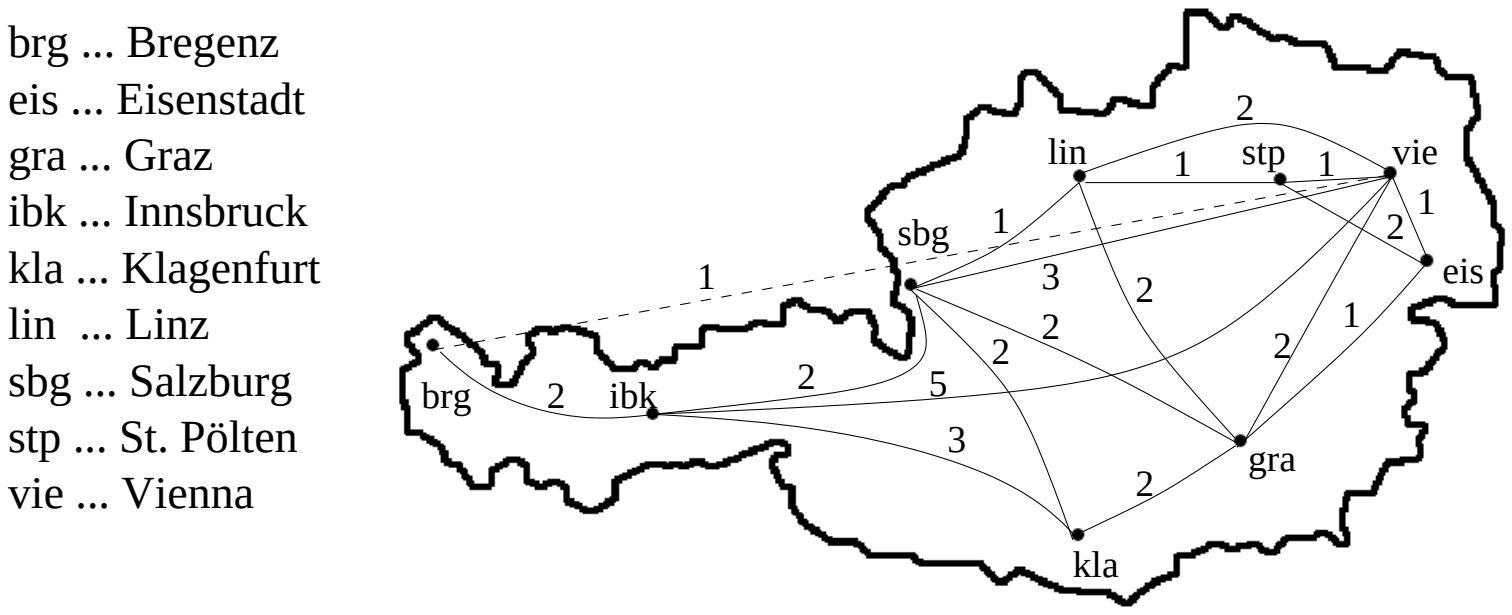

Figure 5: TSP in Austria

We represent this in $\mathcal{K}^c$ as follows. The background knowledge $\Pi_{TSP}$ defines two predicates city(C) and conn(F, T, C) representing the cities and their connections with associated costs. Connections can be traveled in both ways:

```
conn(brg, ibk, 2). conn(ibk, sbg, 2). conn(ibk, vie, 5). conn(ibk, kla, 3).
conn(sbg, kla, 2). conn(sbg, gra, 2). conn(sbg, lin, 1). conn(sbg, vie, 3).
conn(kla, gra, 2). conn(lin, stp, 1). conn(lin, vie, 2). conn(lin, gra, 2).
conn(gra, vie, 2). conn(gra, eis, 1). conn(stp, vie, 1). conn(eis, vie, 1).
conn(stp, eis, 2). conn(vie, brg, 1).
conn(B, A, C) :- conn(A, B, C).
city(T) :- conn(T, _, _).
```

A possible encoding of TSP starting in Vienna (vie) is the $\mathcal{K}^c$ program in Figure 6. It includes two actions for traveling from one city to another and for directly returning to the starting point at the end of the round trip as soon as all cities have been visited.

```
actions :    travel(X, Y) requires conn(X, Y, C) costs C.
             return_from(X) requires conn(X, vie, C) costs C.

fluents :    unvisited. end.
             in(C) requires city(C).
             visited(C) requires city(C).

always :     executable travel(X, Y) if in(X).
             nonexecutable travel(X, Y) if visited(Y).
             executable return_from(X) if in(X).
             nonexecutable return_from(X) if unvisited.

             caused unvisited if city(C), not visited(C).
             caused end after return_from(X).
             caused in(Y) after travel(X, Y).
             caused visited(C) if in(C).
             inertial visited(C).

noConcurrency.

initially :  in(vie).
goal :       end? (9)
```

Figure 6: Traveling Salesperson

The problem has ten optimal 9-step solutions with cost 15. We show only the first five here, as the others are symmetrical:

$P_1 = \langle$ {travel(vie, stp)}, {travel(stp, eis)}, {travel(eis, gra)}, {travel(gra, lin)},
{travel(lin, sbg)}, {travel(sbg, kla)}, {travel(kla, ibk)}, {travel(ibk, brg)},
{return_from(brg)} $\rangle$

$P_2 = \langle$ {travel(vie, eis)}, {travel(eis, stp)}, {travel(stp, lin)}, {travel(lin, sbg)},
{travel(sbg, gra)}, {travel(gra, kla)}, {travel(kla, ibk)}, {travel(ibk, brg)},
{return_from(brg)} $\rangle$

$P_3 = \langle$ {travel(vie, eis)}, {travel(eis, stp)}, {travel(stp, lin)}, {travel(lin, gra)},
{travel(gra, kla)}, {travel(kla, sbg)}, {travel(sbg, ibk)}, {travel(ibk, brg)},
{return_from(brg)} $\rangle$

$P_4 = \langle$ {travel(vie, lin)}, {travel(lin, stp)}, {travel(stp, eis)}, {travel(eis, gra)},
{travel(gra, kla)}, {travel(kla, sbg)}, {travel(sbg, ibk)}, {travel(ibk, brg)},

$\{\texttt{return_from(brg)}\}\ \rangle$

$P_5 = \langle\,\{\texttt{travel(vie,gra)}\},\ \{\texttt{travel(gra,eis)}\},\ \{\texttt{travel(eis,stp)}\},\ \{\texttt{travel(stp,lin)}\},$
$\{\texttt{travel(lin,sbg)}\},\ \{\texttt{travel(sbg,kla)}\},\ \{\texttt{travel(kla,ibk)}\},\ \{\texttt{travel(ibk,brg)}\},$
$\{\texttt{return_from(brg)}\}\ \rangle$

**TSP with variable costs.** Let us now consider an elaboration of TSP, where we assume that the costs of traveling different connections may change during the trip. Note that three of the five solutions in our example above include traveling from St.Pölten to Eisenstadt or vice versa on the second day. Let us now assume that the salesperson, who starts on Monday, has to face some exceptions which might increase the cost of the trip. For instance, (i) heavy traffic jams are expected on Tuesdays on the route from St.Pölten to Eisenstadt or (ii) the salesperson shall not use the flight connection between Vienna and Bregenz on Mondays as only expensive business class tickets are available on this connection in the beginning of the week. So we have to deal with different costs for the respective connections depending on the particular day.

To this end, we first add to the background knowledge $\Pi_{TSP}$ a new predicate cost(A, B, W, C) representing the cost C of traveling connection A to B on weekday W which can take exceptional costs into account:

```
cost(A,B,W,C) :- conn(A,B,C), #int(W), 0 < W, W <= 7, not ecost(A,B,W).
ecost(A,B,W)  :- conn(A,B,C), cost(A,B,W,C1), C != C1.
```

The original costs in the predicate conn(A, B, C) now represent defaults, which can be overridden by explicitly adding different costs. For instance, to represent the exceptions (i) and (ii), we add:

```
cost(stp,eis,2,10).  cost(vie,brg,1,10).
```

setting the exceptional costs for these two critical connections to 10. Weekdays are coded by integers from 1 (Monday) to 7 (Sunday). We represent a mapping from time steps to the weekdays by the following rules which we also add to $\Pi_{TSP}$:

```
weekday(1,1).
weekday(D,W) :- D = D1 + 1, W = W1 + 1, weekday(D1,W1), W1 < 7.
weekday(D,1) :- D = D1 + 1, weekday(D1,7).
```

Note that although the modified background knowledge $\Pi_{TSP}$ is not stratified (since cost is defined by cyclic negation), it has a total well-founded model, and thus a unique answer set.

Finally, we change the costs of traveling and returning in the $\mathcal{K}^c$ program from Figure 6:

```
actions:      travel(X,Y) requires conn(X,Y,C1) costs C
                          where weekday(time,W), cost(X,Y,W,C).
         return_from(X) requires conn(X,vie,C1) costs C
                          where weekday(time,W), cost(X,vie,W,C).
```

Since now the costs for $P_1$ (which includes traveling from St.Pölten to Eisenstadt) on the second day have increased due to exception (i), only four of the plans from above remain optimal. Note that unlike the default costs, exceptional costs do not apply bidirectionally, so the exception does not affect $P_2$ and $P_3$. Furthermore, due to exception (ii) the symmetrical round trips starting with the flight trips to Bregenz are no longer optimal.

The presented encoding proves to be very flexible, as it allows for adding arbitrary exceptions for any connection on any weekday by simply adding the respective facts; moreover, even more involved scenarios, where exceptions are defined by rules, can be modeled.

### 5.3 A Small Example for Planning under Resource Restrictions

Although planning with resources is not the main target of our approach, the following encoding shows that action costs can also be used in order to model optimization of resource consumption in some cases. An important resource in real world planning is money. For instance, let us consider a problem about buying and selling (Lee & Lifschitz, 2001):

> "I have $6 in my pocket. A newspaper costs $1 and a magazine costs $3. Do I have enough money to buy one newspaper and two magazines?"

In $\mathcal{K}^c$, this can be encoded in a very compact way by the following background facts:

$$\texttt{item(newspaper,1).}\quad \texttt{item(magazine,2).}$$

combined with the following short $\mathcal{K}^c$ program:

```
actions:  buy(Item, Number) requires item(Item, Price), #int(Number)
                            costs C where C = Number * Price.

fluents:  have(Item, Number) requires item(Item, Price), #int(Number).

always:   executable buy(Item, Number).
          nonexecutable buy(Item, N1) if buy(Item, N2), N1 < N2.
          caused have(Item, Number) after buy(Item, Number).

goal:     have(newspaper, 1), have(magazines, 2) ? (1)
```

The action buy is always executable, but one must not buy two different amounts of a certain item at once. Obviously, no admissible plan wrt. cost 6 exists, as the optimal plan for this problem, $\langle \{\texttt{buy(newspaper,1)}, \texttt{buy(magazine,2)}\} \rangle$ has $cost^*_{\mathcal{P}} = 7$. Therefore, the answer to the problem is "no."

Our approach considers only positive action costs and does not directly allow modeling full consumer/producer/provider relations on resources in general, in favor of a clear non-ambiguous definition of optimality. For instance, by allowing negative costs one could always add a producer action to make an existing plan cheaper, whereas in our approach costs are guaranteed to increase monotonically, allowing for a clear definition of plan costs and optimality.

On the other hand, we can encode various kinds of resource restrictions by using fluents to represent these resources. We can then model production/consumption as action effects on these fluents and add restrictions as constraints. This allows us to model even complex resource or scheduling problems; optimization, however, remains restricted to action costs.

## 6. Transformation to Logic Programming

In this section, we describe how planning under action costs can be implemented by means of a transformation to answer set programming. It extends our previous transformation (Eiter et al., 2003a), which maps ordinary $\mathcal{K}$ planning problems to disjunctive logic programs under the answer set semantics (Gelfond & Lifschitz, 1991), and takes advantage of weak constraints, cf. (Buccafurri, Leone, & Rullo, 1997, 2000), as implemented in the DLV system (Faber & Pfeifer, 1996; Eiter, Faber, Leone, & Pfeifer, 2000a). In addition, we show how this translation can be adapted to the language of Smodels (Simons, Niemelä, & Soininen, 2002).

### 6.1 Disjunctive Logic Programs with Weak Constraints

First, we give a brief review of disjunctive logic programs with weak constraints.

**Syntax** A *disjunctive rule* (for short, *rule*) $R$ is a construct

$$a_1 \ \mathtt{v} \ \cdots \ \mathtt{v} \ a_n \ \mathtt{:-} \ b_1, \cdots, b_k, \ \mathtt{not}\ b_{k+1}, \cdots, \ \mathtt{not}\ b_m. \tag{2}$$

where all $a_i$ and $b_j$ are classical literals over a function-free first-order alphabet, and $n \geq 0$, $m \geq k \geq 0$. The part left (resp. right) of "$\mathtt{:-}$" is the *head* (resp. *body*) of $R$, where "$\mathtt{:-}$" is omitted if $m = 0$. We let $H(R) = \{a_1, \ldots, a_n\}$ be the set of head literals and $B(R) = B^+(R) \cup B^-(R)$ the set of body literals, where $B^+(R) = \{b_1, \ldots, b_k\}$ and $B^-(R) = \{b_{k+1}, \ldots, b_m\}$. A *(strong) constraint* is a rule with empty head ($n = 0$).

A *weak constraint* is a construct

$$:\sim \ b_1, \cdots, b_k, \ \mathtt{not}\ b_{k+1}, \cdots, \ \mathtt{not}\ b_m. \ [w:] \tag{3}$$

where $w$ is an integer constant or a variable occurring in $b_1, \ldots, b_k$ and all $b_i$ are classical literals.[6] $B(R)$ is defined as for (2).

A *disjunctive logic program (DLP$^w$)* (simply, *program*) is a finite set of rules, constraints and weak constraints; here, superscript $w$ indicates the potential presence of weak constraints.

**Semantics** The answer sets of a program $\Pi$ without weak constraints are defined as usual (Gelfond & Lifschitz, 1991; Lifschitz, 1996). There is one difference, though: We do not consider inconsistent answer sets. The answer sets of a program $\Pi$ with weak constraints are defined by selection from the answer sets $S$ of the weak-constraint free part $\Pi'$ of $\Pi$ as optimal answer sets.

A weak constraint $c$ of form (3) is violated, if it has an instance for which its conjunction is satisfied with respect to the candidate answer set $S$, i.e., there exists a substitution mapping $\theta$ from the variables in $c$ to the Herbrand base of $\Pi$ such that $\{b_1\theta, \cdots, b_k\theta\} \subseteq S$ and $\{b_{k+1}\theta, \cdots, b_m\theta\} \cap M = \emptyset$; we then call $w\theta$ the *violation value* of $c$ wrt. $\theta$.[7] The *violation cost* of $c$ wrt. $S$, denoted $\mathsf{cost}_c(S)$, is the sum of all violation values over all violating substitutions for $c$ wrt. $S$; the *cost of* $S$, denoted $\mathsf{cost}_\Pi(S)$, is then

$$\mathsf{cost}_\Pi(S) = \sum_{c \,\in\, \text{weak constraints of } \Pi} \mathsf{cost}_c(S),$$

i.e., the sum of violation costs of weak constraints in $\Pi$ wrt. $S$. An answer set $M$ of $\Pi$ is now selected (called an *optimal answer set*), if $\mathsf{cost}_\Pi(M)$ is minimal over all answer sets of $\Pi$.

From (Buccafurri et al., 2000) we know that given a head-cycle-free disjunctive program, deciding whether a query $q$ is true in some optimal answer set is $\Delta_2^P$-complete. The respective class for computing such an answer set is $\mathrm{F}\Delta_2^P$-complete. Together with the results from Section 4 this indicates that translations of optimal planning problems to head-cycle-free disjunctive logic programs with weak constraints or the language of Smodels are feasible in polynomial time.

### 6.2 Translating $\mathcal{K}^c$ to DLP$^w$

We extend our original transformation $lp(\mathcal{P})$, which naturally maps a $\mathcal{K}$ planning problem $\mathcal{P}$ into a weak-constraint free program (Eiter et al., 2003a), to a new translation $lp^w(\mathcal{P})$, such that the optimal answer sets of $lp^w(\mathcal{P})$ correspond to the optimal cost plans for the $\mathcal{K}^c$ planning problem $\mathcal{P}$.

6. The colon in $[w:]$ stems from the DLV language, which allows to specify a priority layer after the colon. We do not need priority layers in our translation, but stick to the DLV syntax.
7. A weak constraint $c$ is only admissible, if all possible violation values in all candidate answer sets $S$ are integers. Thus, if $w$ is a variable, then $\Pi$ must guarantee that $w$ can only be bound to an integer.

Basically, in $lp(\mathcal{P})$ fluent and action literals are extended by an additional time parameter, and executability conditions as well as causations rules are modularly translated (rule by rule) into corresponding program rules and constraints; disjunction is used for guessing the actions which should be executed in the plan at each point in time.

#### 6.2.1 REVIEW OF THE TRANSLATION $lp(\mathcal{P})$

The basic steps of the translation from $\mathcal{K}$ programs to logic programs are as follows (cf. (Eiter et al., 2003a) for details):

**Step 0 (Macro Expansion):** First, replace all macros in the $\mathcal{K}$ program by their definitions.

**Step 1 (Background Knowledge):** The background knowledge $\Pi$ of $\mathcal{P}$ is already given as a logic program and is included in $lp(\mathcal{P})$, without further modification.

**Step 2 (Auxiliary Predicates):** To represent steps, we add the following facts to $lp(\mathcal{P})$

$\mathtt{time(0).}, \ldots, \mathtt{time}(l).\quad \mathtt{next(0,1).}, \ldots, \mathtt{next}(l-1, l).$

where $l$ is the plan length of the query $q = G?(l)$ in $\mathcal{P}$ at hand.

**Step 3 (Causation Rules):** Causation rules are mapped to rules in $lp(\mathcal{P})$ by adding type information and extending fluents and actions with a time stamp using `time` and `next`. For example,

`caused across(X) after cross(X), -across(X).`

leads to rule `across(X,T1) :- cross(X,T0), -across(X,T0), person(X), next(T0,T1).`
in $lp(\mathcal{P})$ where $\mathtt{T_1}, \mathtt{T_0}$ are new variables. Here, type information `person(X)` for `across(X)`, and `-across(X)`, taken from the type declaration, is added, which helps to avoid unsafe logic programming rules.

**Step 4 (Executability Conditions):** Similarly, each executability condition is translated to a disjunctive rule “guessing” whether an action occurs at a certain time step. In our running example,

`executable cross(X) if hasLamp(X).`

becomes $\mathtt{cross(X,T_0)} \lor \mathtt{-cross(X,T_0)}$ `:- hasLamp(X,T0), person(X), next(T0,T1).`
which encodes a guess whether at time point $\mathtt{T_0}$ action `cross(X)` should happen; again, type information `person(X)` is added as well as $\mathtt{next(T_0,T_1)}$ to ensure that $\mathtt{T_0}$ is not the last time point.

**Step 5 (Initial State Constraints):** Initial state constraints are transformed like static causation rules in Step 3, but using the constant 0 instead of the variable $\mathtt{T_1}$ and thus need no auxiliary predicate for the time stamp. For instance,

`initially : caused -across(X).`

becomes, by again adding the type information `-across(X,0) :- person(X).`

**Step 6 (Goal Query):** Finally, the query $q$:

$\mathtt{goal:}\ g_1(\overline{t_1}), \ldots, g_m(\overline{t_m}), \mathtt{not}\ g_{m+1}(\overline{t_{m+1}}), \ldots, \mathtt{not}\ g_n(\overline{t_n})\ ?\ (l).$

is translated as follows, where `goal_reached` is a new 0-ary predicate symbol:

$\mathtt{goal_reached\ :-}\ g_1(\overline{t_1}, l), \ldots, g_m(\overline{t_m}, l), \mathtt{not}\ g_{m+1}(\overline{t_{m+1}}, l), \ldots, \mathtt{not}\ g_n(\overline{t_n}, l).$
`:- not goal_reached.`

#### 6.2.2 EXTENDING THE TRANSLATION TO ACTION COSTS

The extended translation $lp^w(\mathcal{P})$ for a $\mathcal{K}^c$ problem $\mathcal{P}$ first includes all rules of $lp(\mathcal{P}_{nc})$, where $\mathcal{P}_{nc}$ results from $\mathcal{P}$ by stripping off all cost parts. Furthermore, the following step is added:

**Step 7 (Action Costs):** For any action declaration $d$ of form (1) with a nonempty `costs`-part, add:

(i) A new rule $r_d$ of the form

$$\mathtt{cost_p(X_1,\ldots,X_n,T,C\theta)} \text{ :- } \mathtt{p(X_1,\ldots,X_n,T),\ t_1,\ldots,t_m,\ c_1\theta,\ldots,c_k\theta, U = T + 1.} \quad (4)$$

where $\mathtt{cost_p}$ is a new symbol, $\mathtt{T}$ and $\mathtt{U}$ are new variables and $\theta = \{\mathtt{time} \to \mathtt{U}\}$. As an optimization, $\mathtt{U} = \mathtt{T} + 1$ is only present if $\mathtt{U}$ occurs elsewhere in $r_d$.

(ii) A weak constraint $wc_d$ of the form

$$:\sim\ \mathtt{cost_p(X_1,\ldots,X_n,T,C).\ [C:]} \quad (5)$$

For example, the `cross` action from the Quick Bridge Crossing Problem is translated to

$\mathtt{cost_{cross}(X,T,WX)}$ :- $\mathtt{cross(X,T),\ person(X),\ walk(X,WX).}$

$:\sim\ \mathtt{cost_{cross}(X,T,WX).\ [WX:]}$

As we showed in previous work (Eiter et al., 2003a), the answer sets of $lp(\mathcal{P})$ correspond to trajectories of optimistic plans for $\mathcal{P}$. The following theorem states a similar correspondence result for $lp^w(\mathcal{P})$ and optimal plans for $\mathcal{P}$. We define, for any consistent set of ground literals $S$, the sets $A_j^S = \{a(\bar{t}) \mid a(\bar{t}, j-1) \in S,\ a \in \sigma^{act}\}$ and $s_j^S = \{f(\bar{t}) \mid f(\bar{t}, j) \in S,\ f(\bar{t}) \in \mathcal{L}_{fl}\}$, for all $j \geq 0$.

**Theorem 6.1 (Answer Set Correspondence)** *Let $\mathcal{P} = \langle PD, q\rangle$ be a (well-defined) $\mathcal{K}^c$ planning problem, and let $lp^w(\mathcal{P})$ be the above program. Then,*

*(i) for each optimistic plan $P = \langle A_1,\ldots,A_l\rangle$ of $\mathcal{P}$ and supporting trajectory $T = \langle\langle s_0, A_1, s_1\rangle, \langle s_1, A_2, s_2\rangle, \ldots, \langle s_{l-1}, A_l, s_l\rangle\rangle$ of $P$, there exists some answer set $S$ of $lp^w(\mathcal{P})$ such that $A_j = A_j^S$ for all $j = 1,\ldots,l$, $s_j = s_j^S$, for all $j = 0,\ldots,l$ and $cost_{\mathcal{P}}(P) = \mathsf{cost}_{lp^w(\mathcal{P})}(S)$;*

*(ii) for each answer set $S$ of $lp^w(\mathcal{P})$, the sequence $P = \langle A_1,\ldots,A_l\rangle$ is a solution of $\mathcal{P}$, i.e., an optimistic plan, witnessed by the trajectory $T = \langle\langle s_0, A_1, s_1\rangle, \langle s_1, A_2, s_2\rangle, \ldots, \langle s_{l-1}, A_l, s_l\rangle\rangle$ with $cost_{\mathcal{P}}(P) = \mathsf{cost}_{lp^w(\mathcal{P})}(S)$, where $A_j = A_j^S$ and $s_k = s_k^S$ for all $j = 1,\ldots,l$ and $k = 0,\ldots,l$.*

The proof is based on the resp. correspondence result for $\mathcal{K}$ (Eiter et al., 2003a). For the details, we refer to the Appendix.

From this result and the definitions of optimal cost plans and optimal answer sets, we conclude the following result:

**Corollary 6.2 (Optimal answer set correspondence)** *For any well-defined $\mathcal{K}^c$ planning problem $\mathcal{P} = \langle PD, Q\,?\,(l)\rangle$, the trajectories $T = \langle\langle s_0, A_1, s_1\rangle, \ldots, \langle s_{l-1}, A_l, s_l\rangle\rangle$ of optimal plans $P$ for $\mathcal{P}$ correspond to the optimal answer sets $S$ of $lp^w(\mathcal{P})$, such that $A_j = A_j^S$ for all $j = 1,\ldots,l$ and $s_j = s_j^S$, for all $j = 0,\ldots,l$.*

*Proof.* For each $a \in A_j$, the weak constraint (5) causes a violation value of $cost_j(a)$. Furthermore, these are the only cost violations. Thus, a candidate answer set $S$ is optimal if and only if $\mathsf{cost}_{lp^w(\mathcal{P})}(S) = \sum_{j=1}^{l} \sum_{a \in A_j} cost_j(a) = cost_{\mathcal{P}}(P)$ is minimal, i.e., $S$ corresponds to an optimal plan. □

A similar correspondence result also holds for admissible plans:

**Corollary 6.3 (Answer set correspondence for admissible plans)** *For any well-defined $\mathcal{K}^c$ planning problem $\mathcal{P} = \langle PD, Q\,?\,(l)\rangle$, the trajectories $T = \langle\langle s_0, A_1, s_1\rangle, \ldots, \langle s_{l-1}, A_l, s_l\rangle\rangle$ of admissible plans $P$ for $\mathcal{P}$ wrt. cost $c$ correspond to the answer sets $S$ of $lp^w(\mathcal{P})$ having $\mathsf{cost}_{lp^w(\mathcal{P})}(S) \leq c$, such that $A_j = A_j^S$ for all $j = 1, \ldots, l$ and $s_j = s_j^S$, for all $j = 0, \ldots, l$.*

As for secure planning, we have introduced a technique to check security of an optimistic plan for certain planning problem instances by means of a logic program (Eiter et al., 2003a). This method carries over to planning with action costs in a straightforward way, and optimal resp. admissible secure plans can be similarly computed by answer set programming.

### 6.3 Alternative Translation for Smodels

Apart from the presented translation using weak constraints, one could also choose an alternative approach for the translation to answer set programming. Smodels (Simons et al., 2002) supports another extension to pure answer set programming allowing to minimize over sets of predicates.

This approach could be used in an alternative formulation of Step 7:

**Step 7a:** For action declarations with nonempty `costs`-parts, we add a new rule of form

$$\mathtt{cost}(\mathtt{p}, \mathtt{X_1}, \ldots, \mathtt{X_n}, \mathtt{0}, \ldots, \mathtt{0}, \mathtt{T}, \mathtt{C}\theta) \text{ :- } \mathtt{t_1}, \ldots, \mathtt{t_m}, \mathtt{c_1}\theta, \ldots, \mathtt{c_k}\theta, \mathtt{U} = \mathtt{T} + \mathtt{1}. \qquad (6)$$

similar to Step 7 above, with two differences: (1) action name `p` is now a parameter, and (2) we add $l - \mathtt{n}$ parameters with constant "0" between $\mathtt{X_n}$ and `T` where $l$ is the maximum arity of all actions in $PD$. This is necessary in order to get unique arity $l + 2$ for predicate `cost`. Furthermore, we add

$$\mathtt{occurs}(\mathtt{p}, \mathtt{X_1}, \ldots, \mathtt{X_n}, \mathtt{0}, \ldots, \mathtt{0}, \mathtt{T}) \text{ :- } \mathtt{p}(\mathtt{X_1}, \ldots, \mathtt{X_n}, \mathtt{T}), \mathtt{t_1}, \ldots, \mathtt{t_m},. \qquad (7)$$

This second rule adds the same "0" parameters as for to achieve unique arity $l + 1$ of the new predicate `occurs`. Using Smodels syntax, we can now compute optimal plans by adding

$\mathtt{minimize}[\mathtt{occurs}(\mathtt{A}, \mathtt{X_1}, ..., \mathtt{X}_l, \mathtt{T}) : \mathtt{cost}(\mathtt{A}, \mathtt{X_1}, ..., \mathtt{X}_l, \mathtt{T}, \mathtt{C}) = \mathtt{C}].$

Note that Smodels does not support disjunction in rule heads, so we also need to modify Step 4, expressing the action guess via unstratified negation or Smodels' choice rules.

## 7. Implementation

We have implemented an experimental prototype system, DLV$^{\mathcal{K}}$, for solving $\mathcal{K}$ planning problems (Eiter et al., 2003a). An improved version of this prototype it is now capable of optimal and admissible planning with respect to the extended syntax of $\mathcal{K}^c$, available for experiments at `http://www.dlvsystem.com/K/`.

DLV$^{\mathcal{K}}$ has been realized as a frontend to the DLV system (Faber & Pfeifer, 1996; Eiter et al., 2000a). First, the planning problem at hand is transformed as described in the previous section. Then, the DLV kernel is invoked to produce answer sets. For *optimistic planning* the (optimal, if action costs are defined) answer sets are then simply translated back into suitable output for the user and printed.

In case the user specified that *secure/conformant planning* should be performed, our system has to check security of the plans computed. In normal (non-optimal) planning, this is simply done by checking each answer set returned right before transforming it back to user output. In the case of optimal secure planning, on the other hand, the candidate answer set generation of the DLV kernel has to be "intercepted": The kernel proceeds computing candidate answer sets, returning an answer set with minimal violation cost value, by running through all candidates. Here, in order to generate *optimal secure plans*, the planning frontend interrupts computation, allowing only answer sets which represent secure plans to be considered as candidates.

Checking plan security is done by rewriting the translated program wrt. the candidate answer set/plan in order to verify whether the plan is secure. The rewritten "check program" is tested by a separate invocation of the DLV kernel. As for further details on the system architecture we refer to (Eiter et al., 2003a)

### 7.1 Usage

Suppose the background knowledge and the program depicted in Figure 1 with the cost extensions from Section 3.3 are stored in files `crossing.bk` and `crossing.plan`; then, by invoking the program with the command line

$$\mathtt{dlv -FPcrossing.plancrossing.bk -planlength = 7}$$

we compute all optimal plans solving this problem in seven steps. In the output we find, after a supporting trajectory, the following optimal plan:

```
PLAN : crossTogether(joe, jack) : 2;  cross(joe) : 1;  takeLamp(william);
       crossTogether(william, averell) : 10;  takeLamp(jack);
       cross(jack) : 2;  crossTogether(joe, jack) : 2   COST :  17
```

For each action, its cost is shown after a colon, if it is non-zero. The switch `-planlength=`$i$ can be used to set the plan length; it overrides any plan length given in the `query`-part of the planing problem. Using `-planlength=5`, we get plans with cost 19, as there are no cheaper plans of that length.

The user is then asked whether to perform the optional security check and whether to look for further (optimal) plans, respectively. The switch `-FPsec` can be used instead of `-FP` to obtain secure plans only.

The command line option `-costbound=`$N$ effects the computation of all admissible plans with respect to cost $N$. For example, the resource problem described in Section 5.3 can be solved by the following call to our prototype:

$$\mathtt{dlv -FPbuying.bkbuying.plan -N = 10 -planlength = 1 -costbound = 6}$$

Correctly, no admissible plan is found. When calling the system again without cost bound, the prototype calculates the following optimal cost plan:

```
PLAN : buy(newspaper, 1) : 1,  buy(magazine, 2) : 6   COST :  7
```

The current prototype supports simple bounded integer arithmetics. The option `-N=10` used above sets an upper bound of $N = 10$ for the integers which may be used in a program; the built-in predicate $\#\mathtt{int}$ is true for all integers $0 \ldots N$. Setting $N$ high enough, taking into account the outcome of built-in arithmetic predicates $A = B + C$ and $A = B * C$, is important to get correct results. Further details on the prototype are given on the DLV$^{\mathcal{K}}$ web site at `http://www.dlvsystem.com/K/`.

## 7.2 Experiments

Performance and experimental results for $\text{DLV}^{\mathcal{K}}$ (without action costs and optimal planning) were reported in previous work (Eiter et al., 2003a). In this section, we present some encouraging experimental results for planning with action costs, in particular for parallel Blocks World and TSP. All experiments were performed on a Pentium III 733MHz machine with 256MB of main memory running SuSE Linux 7.2. We set a time limit of 4000 seconds for each tested instance where exceeding this limit is indicated by "-" in the result tables.

Where possible, we also report results for CCALC and CMBP, two other logic-based planning systems whose input languages ($\mathcal{C}+$ resp. $\mathcal{AR}$) have capabilities similar to $\mathcal{K}$ resp. $\mathcal{K}^c$.

**CCALC.** The *Causal Calculator* (*CCALC*) is a model checker for the languages of causal theories (McCain & Turner, 1997). It translates programs in the action language $\mathcal{C}+$ into the language of causal theories which are in turn transformed into SAT problems; these are then solved using a SAT solver (McCain & Turner, 1998). The current version of CCALC uses mChaff (Moskewicz et al., 2001) as its default SAT solver. Minimal length plans are generated iteratively increasing the plan length up to an upper bound. CCALC is written in Prolog. For our tests, we used version 2.04b of CCALC which we obtained from `<URL:http://www.cs.utexas.edu/users/tag/cc/>` and a trial version of SICStus Prolog 3.9.1. We used encodings taken from (Lee & Lifschitz, 2001) for parallel Blocks World adapted for CCALC 2.0. These encodings are included in the current download version of the system. For sequential Blocks World we adapted the encodings by adding the $\mathcal{C}+$ command "`noConcurrency`." which resembles the respective $\mathcal{K}$ command. All results for CCALC include 2.30sec startup time.

**CMBP.** The *Conformant Model Based Planner* (CMBP) (Cimatti & Roveri, 2000) is based on the model checking paradigm and exploits symbolic Boolean function representation techniques such as Binary Decision Diagrams (Bryant, 1986). CMBP allows for computing sequential minimal length plans, where the user has to declare an upper bound for the plan length. Its input language is an extension of $\mathcal{AR}$ (Giunchiglia, Kartha, & Lifschitz, 1997). Unlike $\mathcal{K}$ or action languages such as $\mathcal{C}+$ (Lee & Lifschitz, 2001), this language only supports propositional actions. CMBP is tailored for conformant planning. The results reported complement a previous comparison which also shows the encoding for sequential Blocks World in CMBP (Eiter et al., 2003a). For our tests, we used CMBP 1.0, available at `<URL:http://sra.itc.it/people/roveri/cmbp/>`.

### 7.2.1 BLOCKS WORLD

Tables 1–4 show the results for our different Blocks World encodings in Section 5.1 on several configurations: P0 denotes our simple instance from Figure 3, while P1–P5 are instances used in previous work (Eiter et al., 2003a; Erdem, 1999).

Table 1 shows the results for finding a shortest sequential plan. The second and third column show the number of blocks and the length of a shortest plan (i.e., the least number of moves) solving the respective blocks world instance. The execution time for solving the problem using the shortest-plan encoding $\mathcal{P}_\beta$ in Section 5.1 is shown in column five, using the upper bound shown in the fourth column on the plan length. Column six shows the execution time for finding the shortest plan in an incremental plan length search starting from 0, similar to the method used for CCALC. The remaining two columns show the results for CCALC and CMBP.

| Problem | #blocks | min. #moves (=#steps) | upper bound #steps | $\text{DLV}^{\mathcal{K}}$ | $\text{DLV}^{\mathcal{K}}_{inc}$ | CCALC | CMBP |
|---|---|---|---|---|---|---|---|
| P0 | 6 | 5 | 6 | 0.48s | 0.29s | 4.65s | 21.45s |
| P1 | 4 | 4 | 4 | 0.05s | 0.08s | 3.02s | 0.13s |
| P2 | 5 | 6 | 7 | 0.24s | 0.27s | 4.02s | 8.44s |
| P3 | 8 | 8 | 10 | 25.32s | 2.33s | 10.07s | - |
| P4 | 11 | 9 | 16 | - | 8.28s | 27.19s | - |
| P5 | 11 | 11 | 16 | - | 12.63s | 32.27s | - |

Table 1: Sequential Blocks World - shortest plans

| Problem | #blocks | #steps(fixed) | min. #moves | $\text{DLV}^{\mathcal{K}}$ |
|---|---|---|---|---|
| P0 | 6 | 2 | 6 | 0.05s |
| P0 | 6 | 3 | 5 | 0.09s |
| P1 | 4 | 3 | 4 | 0.04s |
| P2 | 5 | 5 | 6 | 0.10s |
| P3 | 8 | 4 | 9 | 0.21s |
| P4 | 11 | 5 | 13 | 0.81s |
| P5 | 11 | 7 | 15 | 327s |

Table 2: Parallel Blocks World - cheapest plans: Minimal number of moves at fixed plan length ($\alpha$)

Table 2 shows the execution times for parallel blocks world with fixed plan length where the number of moves is minimized, i.e. problem ($\alpha$) in Section 5.1. We used the encoding in Figure 2, which generates parallel serializable plans. As CCALC and CMBP do not allow for optimizing other criteria than plan length, we only have results for $\text{DLV}^{\mathcal{K}}$ here.

Next, Table 3 shows some results for finding a shortest parallel plan, i.e. problem ($\beta$) in Section 5.1. First, the minimal possible number of steps is given. We processed each instance (i) using the encoding $\mathcal{P}_{\beta}$ from Section 5.1, (ii) without costs by iteratively increasing the plan length and (iii) using CCALC, by iteratively increasing the plan length until a plan is found. For every result, the number of moves of the first plan computed is reported separately. As CMBP only supports sequential planning, it is not included in this comparison.

Finally, Table 4 shows the results for the combined optimizations ($\gamma$) and ($\delta$) for parallel Blocks World as outlined in Section 5.1. The second column again contains the upper bound for the plan

| | upper bound | min. #steps | $\text{DLV}^{\mathcal{K}}$ | | $\text{DLV}^{\mathcal{K}}_{inc}$ | | CCALC | |
|---|---|---|---|---|---|---|---|---|
| | | | #moves | time | #moves | time | #moves | time |
| P0 | 6 | 2 | 6 | 0.52s | 6 | 0.09s | 6 | 4.05s |
| P1 | 4 | 3 | 5 | 0.07s | 5 | 0.08s | 4 | 2.95s |
| P2 | 7 | 5 | 9 | 0.39s | 9 | 0.21s | 6 | 3.70s |
| P3 | 10 | 4 | - | - | 12 | 0.43s | 9 | 7.69s |
| P4 | 16 | 5 | - | - | 18 | 1.54s | 13 | 20.45s |
| P5 | 16 | 7 | - | - | 26 | 3.45s | 15 | 23.22s |

Table 3: Parallel Blocks World - shortest plan ($\beta$)

| | upper bound | (γ) | | | | (δ) | | |
|---|---|---|---|---|---|---|---|---|
| | | steps/moves | $\text{DLV}^{\mathcal{K}}$ | $\text{DLV}^{\mathcal{K}}_{inc}$ | CCALC | steps/moves | $\text{DLV}^{\mathcal{K}}$ | $\text{DLV}^{\mathcal{K}}_{inc}$ |
| P0 | 6 | 3/5 | 38.5s | 0.18s | 5.89s | 2/6 | 0.26s | 0.09s |
| P1 | 4 | 3/4 | 0.07s | 0.11s | 3.47s | 3/4 | 0.08s | 0.08s |
| P2 | 7 | 5/6 | 2.08s | 0.21s | 5.65s | 5/6 | 0.78s | 0.28s |
| P3 | 10 | 5/8 | - | 1.57s | 15.73s | 4/9 | 177s | 0.45s |
| P4 | 16 | 9/9 | - | - | 73.64s | 5/13 | - | 1.86s |
| P5 | 16 | 11/11 | - | - | 167s | 7/15 | - | 323s |

Table 4: Parallel Blocks World - ($\gamma$),($\delta$)

length of the respective instance. The following three columns present the results on finding a shortest among the cheapest plans, i.e. problem ($\gamma$) in Section 5.1:

**$\text{DLV}^{\mathcal{K}}$** refers to the results for our combined minimal encoding $\mathcal{P}_\gamma$ and as described in Section 5.1;

**$\text{DLV}^{\mathcal{K}}_{inc}$** refers to the results for incrementally searching for the shortest among the cheapest plans: This is done by means of the `-costbound=`$i$ command line option taking the minimal sequential costs (i.e., the shortest sequential plan length as computed in Table 1) as an upper cost limit. As our encodings compute serializable plans, the minimal sequential length can be used as cost limit in this special case.

**CCALC** A similar technique can be used with CCALC when encoding bound costs through "additive fluents" (Lee & Lifschitz, 2001).

Note that the incremental strategy (used by $\text{DLV}^{\mathcal{K}}_{inc}$ and CCALC) takes advantage of our specific formulation of the parallel Blocks World problem: In general, when allowing parallel actions which are not necessarily serializable and have arbitrary costs, the optimal parallel cost might differ from the optimal sequential solution. In particular, plans which are longer than the cheapest sequential plans (which, in this example, coincide with the shortest sequential plans) may need to be considered. This makes incremental search for a solution of problem ($\gamma$) infeasible in general.

The last test is finding a cheapest among the shortest plans, that is, problem ($\delta$) in Section 5.1. Again we have tested the integrated encoding with an upper bound ($\mathcal{P}_\delta$) resp. incrementally finding the shortest plan. Unlike for problem ($\gamma$), we cannot derive a fixed cost limit from the sequential solution here; we really need to optimize costs, which makes an encoding in CCALC infeasible.

**Blocks World – Results** The Blocks World experiments show that $\text{DLV}^{\mathcal{K}}$ can solve various optimization tasks in a more effective and flexible way than the systems compared. On the other hand, as already stated above, for the minimal plan length encodings in Section 5.1, we can only solve the problems where a tight upper bound for the plan length is known. Iteratively increasing the plan length is more effective, especially if the upper bound is much higher than the actual optimal solution. This becomes drastically apparent when execution times seem to explode from one instance to the next, in a highly non-linear manner as in Table 1 where a solution for P3 can be found in reasonable time whereas P4 and P5 could not be solved within the time limit of 4000 seconds. This observation is also confirmed in the other tables (instance P5 in Table 2, etc.) and is partly explained by the behavior of the underlying DLV system, which is not geared towards plan search, and as a general purpose problem solver uses heuristics which might not work out well in some cases. In particular, during the answer set generation process in DLV, no distinction is made between actions and fluents, which might be useful for planning tasks to control the generation of answer sets resp. plans; this may be part of further investigations.

Interestingly, CCALC finds "better quality" parallel solutions for problem ($\beta$) (cf. Table 3), i.e. solutions with fewer moves, although it is significantly slower than our system on these instances. For the incremental encoding of problem ($\gamma$), CCALC seems even more effective than our system. However, CCALC offers no means of optimization; it allows for admissible but not for optimal planning. This makes our approach more flexible and general. As stated above, we could fortunately exploit the fixed cost bound in this particular example for CCALC, which is not possible in general instances of problem ($\gamma$).

Problem ($\gamma$) is also intuitively harder than simply finding a shortest plan or a cheapest among all shortest plans in general: While these problems can always be solved incrementally, for ($\gamma$) we must consider all plans of all lengths. A longer plan may be cheaper, so we cannot freeze the plan length once a (shortest) plan has been incrementally found.

#### 7.2.2 TSP

Some experimental results on TSP with variable costs are reported in Tables 5 and 6. Unlike for blocks world, no comparable systems were available; none of the systems from above supports cost optimal planning as needed for solving this problem. Here, the plan length is always given by the number of cities.

Table 5 shows the results for our TSP instance on the Austrian province capitals as in Figure 5 (nine cities, 18 connections), with and without the exceptional costs as in Section 5.2 (with and without subscript $exc$ in the table). Further instances reported in this table with different cost exceptions ($we$, $lwe$, $rnd$) are described below.

Results for some bigger TSP instances, given by the capitals of the 15 members of the European Union (EU) with varying connection graphs and exceptional costs are shown in Table 6. We have used the flight distances (km) between the cities as connection costs. Instances TSP$_{EU1}$–TSP$_{EU6}$ have been generated by randomly choosing a given number of connections from all possible connections between the 15 cities. Note that TSP$_{EU1}$ has no solution; the time reported here is until DLV$^{\mathcal{K}}$ terminated, and for all other instances until the first optimal plan was found.

We have also tested some instances of more practical relevance than simply randomly choosing connections: TSP$_{EU7}$ is an instance where we have taken the flight connections of three carriers (namely, Star Alliance, Alitalia, and Luxair), and in TSP$_{EU8}$ we have included only direct connections of at most 1500km. Such a "capital hopping" is of interest for a small airplane with limited range, for instance.

For each instance in Tables 5–6 we have measured the execution time:

- without exceptional costs,
- with 50% costs for all connections on Saturdays and Sundays (weekends, $we$)
- with 50% costs for all connections on Fridays, Saturdays and Sundays (long weekends, $lwe$),
- for some random cost exceptions ($rnd$): We have added a number of randomly generated exceptions with costs between 0 and 10 for TSP$_{Austria}$ and between 0 and 3000 for the instances *EU1* to *EU8*.

| Instance | #cost exceptions | cost/time |
|---|---|---|
| $\text{TSP}_{Austria}$ | 0 | 15/0.31s |
| $\text{TSP}_{Austria,exc}$ | 2 | 15/0.32s |
| $\text{TSP}_{Austria,we}$ | 36 | 12/0.34s |
| $\text{TSP}_{Austria,lwe}$ | 54 | 11/0.35s |
| $\text{TSP}_{Austria,rnd}$ | 10 | 14/0.30s |
| $\text{TSP}_{Austria,rnd}$ | 50 | 15/0.31s |
| $\text{TSP}_{Austria,rnd}$ | 100 | 23/0.35s |
| $\text{TSP}_{Austria,rnd}$ | 200 | 36/0.37s |

Table 5: TSP – Results for $\text{TSP}_{Austria}$ with varying exceptions

| Instance | #conn. | #except. | cost/time |
|---|---|---|---|
| $\text{TSP}_{EU1}$ | 30 | 0 | -/9.11s |
| $\text{TSP}_{EU1,we}$ | 30 | 60 | -/11.93s |
| $\text{TSP}_{EU1,lwe}$ | 30 | 90 | -/13.82s |
| $\text{TSP}_{EU1,rnd}$ | 30 | 100 | -/11.52s |
| $\text{TSP}_{EU1,rnd}$ | 30 | 200 | -/12.79s |
| $\text{TSP}_{EU1,rnd}$ | 30 | 300 | -/14.64s |
| $\text{TSP}_{EU1,rnd}$ | 30 | 400 | -/16.26s |
| $\text{TSP}_{EU2}$ | 30 | 0 | 16213/13.27s |
| $\text{TSP}_{EU2,we}$ | 30 | 60 | 13195/16.41s |
| $\text{TSP}_{EU2,lwe}$ | 30 | 90 | 11738/18.53s |
| $\text{TSP}_{EU2,rnd}$ | 30 | 100 | 15190/15.54s |
| $\text{TSP}_{EU2,rnd}$ | 30 | 200 | 13433/16.31s |
| $\text{TSP}_{EU2,rnd}$ | 30 | 300 | 13829/18.34s |
| $\text{TSP}_{EU2,rnd}$ | 30 | 400 | 13895/20.59s |
| $\text{TSP}_{EU3}$ | 35 | 0 | 18576/24.11s |
| $\text{TSP}_{EU3,we}$ | 35 | 70 | 15689/28.02s |
| $\text{TSP}_{EU3,lwe}$ | 35 | 105 | 14589/30.39s |
| $\text{TSP}_{EU3,rnd}$ | 35 | 100 | 19410/26.75s |
| $\text{TSP}_{EU3,rnd}$ | 35 | 200 | 22055/29.64s |
| $\text{TSP}_{EU3,rnd}$ | 35 | 300 | 18354/31.54s |
| $\text{TSP}_{EU3,rnd}$ | 35 | 400 | 17285/32.66s |
| $\text{TSP}_{EU4}$ | 35 | 0 | 16533/36.63s |
| $\text{TSP}_{EU4,we}$ | 35 | 70 | 12747/41.72s |
| $\text{TSP}_{EU4,lwe}$ | 35 | 105 | 11812/43.12s |
| $\text{TSP}_{EU4,rnd}$ | 35 | 100 | 15553/39.17s |
| $\text{TSP}_{EU4,rnd}$ | 35 | 200 | 13216/41.19s |
| $\text{TSP}_{EU4,rnd}$ | 35 | 300 | 16413/43.51s |
| $\text{TSP}_{EU4,rnd}$ | 35 | 400 | 13782/45.69s |
| $\text{TSP}_{EU5}$ | 40 | 0 | 15716/91.83s |
| $\text{TSP}_{EU5,we}$ | 40 | 80 | 12875/97.73s |
| $\text{TSP}_{EU5,lwe}$ | 40 | 120 | 12009/100.14s |
| $\text{TSP}_{EU5,rnd}$ | 40 | 100 | 13146/85.69s |
| $\text{TSP}_{EU5,rnd}$ | 40 | 200 | 12162/83.44s |
| $\text{TSP}_{EU5,rnd}$ | 40 | 300 | 12074/76.81s |
| $\text{TSP}_{EU5,rnd}$ | 40 | 400 | 12226/82.97s |
| $\text{TSP}_{EU5,rnd}$ | 40 | 500 | 13212/82.53s |

| Instance | #conn. | #except. | cost/time |
|---|---|---|---|
| $\text{TSP}_{EU6}$ | 40 | 0 | 17483/142.7s |
| $\text{TSP}_{EU6,we}$ | 40 | 80 | 14336/150.3s |
| $\text{TSP}_{EU6,lwe}$ | 40 | 120 | 13244/154.7s |
| $\text{TSP}_{EU6,rnd}$ | 40 | 100 | 15630/142.5s |
| $\text{TSP}_{EU6,rnd}$ | 40 | 200 | 14258/137.2s |
| $\text{TSP}_{EU6,rnd}$ | 40 | 300 | 11754/120.5s |
| $\text{TSP}_{EU6,rnd}$ | 40 | 400 | 11695/111.4s |
| $\text{TSP}_{EU6,rnd}$ | 40 | 500 | 12976/120.8s |
| $\text{TSP}_{EU7}$ | 55 | 0 | 15022/102.6s |
| $\text{TSP}_{EU7,we}$ | 55 | 110 | 12917/112.2s |
| $\text{TSP}_{EU7,lwe}$ | 55 | 165 | 11498/116.2s |
| $\text{TSP}_{EU7,rnd}$ | 55 | 100 | 13990/104.2s |
| $\text{TSP}_{EU7,rnd}$ | 55 | 200 | 12461/100.8s |
| $\text{TSP}_{EU7,rnd}$ | 55 | 300 | 13838/106.9s |
| $\text{TSP}_{EU7,rnd}$ | 55 | 400 | 12251/96.58s |
| $\text{TSP}_{EU7,rnd}$ | 55 | 500 | 16103/109.2s |
| $\text{TSP}_{EU7,rnd}$ | 55 | 600 | 14890/110.3s |
| $\text{TSP}_{EU7,rnd}$ | 55 | 700 | 17070/110.7s |
| $\text{TSP}_{EU8}$ | 64 | 0 | 10858/3872s |
| $\text{TSP}_{EU8,we}$ | 64 | 128 | 9035/3685s |
| $\text{TSP}_{EU8,lwe}$ | 64 | 192 | 8340/3324s |
| $\text{TSP}_{EU8,rnd}$ | 64 | 100 | 10283/2603s |
| $\text{TSP}_{EU8,rnd}$ | 64 | 200 | 9926/1372s |
| $\text{TSP}_{EU8,rnd}$ | 64 | 300 | 10028/1621s |
| $\text{TSP}_{EU8,rnd}$ | 64 | 400 | 8133/597.7s |
| $\text{TSP}_{EU8,rnd}$ | 64 | 500 | 8770/573.3s |
| $\text{TSP}_{EU8,rnd}$ | 64 | 600 | 8220/360.7s |
| $\text{TSP}_{EU8,rnd}$ | 64 | 700 | 6787/324.6s |
| $\text{TSP}_{EU8,rnd}$ | 64 | 800 | 11597/509.5s |

Table 6: TSP – Various instances for the capitals of the 15 EU members

**TSP – Results** Instance TSP$_{EU8}$ shows the limits of our system: the given data allows for many possible tours, so finding an optimal one gets very tricky. On the other hand, a realistic instance like TSP$_{EU7}$ with real airline connections is solved rather quickly, which is not very surprising: Most airlines have a central airport (for instance Vienna for Austrian Airlines) and few direct connections between the destinations served. This allows for much fewer candidate answer sets, when (as in reality) the number of airlines we consider is limited. E.g., TSP$_{EU7}$ has no solution at all if only two out of Star Alliance, Alitalia, and Luxair are allowed. Of course, we cannot compete with dedicated TSP solvers/algorithms, which are able to solve much bigger TSP instances and have not been considered here. However, to our knowledge, none of these solvers can deal with features such as incomplete knowledge, defaults, time dependent exceptional costs, etc. directly. Our results even show that execution times are stable yet in case of many exceptions. In contrast, instance TSP$_{EU8}$ shows that exceptions can also cause a significant speedup. This is due to the heuristics used by the underlying DLV system, which can single out better solutions faster if costs are not spread evenly like in TSP$_{EU8}$ without exceptional costs.

Note that, we have also experimented with the alternative Smodels translation sketched in Section 6.3. We refrain from detailed discussion here, since the (i) translation is optimized for DLV and Smodels performance was worse (around factor 10 for the tested TSP instances) than DLV and (ii) there is no integrated planning frontend available for Smodels providing a high-level planning language. Nevertheless, we have shown that our approach can, with minor modifications, be adopted in a planning system based on Smodels.

## 8. Related Work

In the last years, it has been widely recognized that plan length alone is only one criterion to be optimized in planning. Several attempts have been made to extend planners to also consider action costs.

The PYRRHUS system (Williams & Hanks, 1994) is an extension of UCPOP planning which allows for optimal planning with resources and durations. Domain-dependent knowledge can be added to direct the heuristic search. A "utility model" has to be defined for a planning problem which can be used to express an optimization function. This system supports a language extension of ADL (Pednault, 1989), which is a predecessor of PDDL (Ghallab et al., 1998). The algorithm is a synthesis of branch-and-bound optimization with a least-commitment, plan-space planner.

Other approaches based on heuristic search include the use of an A* strategy together with action costs in the heuristics (Ephrati, Pollack, & Mihlstein, 1996) and work by Refanidis and Vlahavas who use multi-criteria heuristics to obtain near-optimal plans, considering multiple criteria apart from plan length alone (Refanidis & Vlahavas, 2001). However, the described heuristics is not fully admissible, and only guarantees optimal plans under certain restrictions (Haslum & Geffner, 2000). In fact, most heuristic state-space planners are not able to guarantee optimality.

A powerful approach has been suggested by Nareyek, who describes planning with resources as a structural constraint satisfaction problem (SCSP), and then solves that problem by local search combined with global control. However, this work promotes the inclusion of domain-dependent knowledge; the general problem has an unlimited search space, and no declarative high-level language is provided (Nareyek, 2001).

Among other related approaches, Kautz and Walser generalize the "Planning as Satisfiability" approach to use integer optimization techniques for encoding optimal planning under resource pro-duction/consumption (Kautz & Walser, 1999). First, they recall that integer logic programming generalizes SAT, as a SAT formula can be translated to a system of inequalities. Second, they extend effects and preconditions of actions similar to a STRIPS extension proposed by Koehler for modeling resource consumption/production (Koehler, 1998). Kautz and Walser allow for arbitrary optimization functions but they use a non-declarative, low-level representation based on the algebraic modeling language AMPL (Fourer, Gay, & Kernighan, 1993). They mention that Koehler's STRIPS-like formalization can be mapped to their approach. However, they can not express non-determinism or incomplete knowledge. There is an implementation of this approach called ILP-PLAN, which uses the AMPL package (`http://www.ampl.com/`). Unfortunately, AMPL is not freely available, so we could not compare the system with our approach experimentally.

Lee and Lifschitz describe the extension $\mathcal{C}+$ of the action language $\mathcal{C}$ which allows for an intuitive encoding of resources and costs by means of so called "additive fluents" (Lee & Lifschitz, 2001). This way admissible planning can be realized, but optimization has not been considered in that framework so far. An implementation of a planner based on this language is CCALC (McCain, 1999) which has already been described in the previous section. Another implementation of a planning system based on the action language $\mathcal{C}$ is Cplan (Giunchiglia, 2000; Ferraris & Giunchiglia, 2000). The Cplan system mainly focuses on conformant planning and does not support the advanced features of $\mathcal{C}+$. Furthermore, the code is no longer maintained.

Son and Pontelli propose to translate action language $\mathcal{B}$ to prioritized default theory and answer set programming. They allow to express preferences between actions and rules at the object level in an interpreter but not as a part of the input language (Son & Pontelli, 2002). However, these preferences are orthogonal to our approach as they model qualitative preferences as opposed to our overall value function of plans/trajectories.

## 9. Conclusion and Outlook

This work continues a research stream which pursues the usage of answer set programming for building planning systems which offer declarative planning languages based on action languages, where planning tasks are specified at a high level of abstraction (Lifschitz, 1999a, 1999b). For representation of practical planning problems, such languages must have high expressiveness and provide convenient constructs and language elements.

Towards this goal, we have presented the planning language $\mathcal{K}^c$, which extends the declarative planning language $\mathcal{K}$ (Eiter et al., 2000b, 2003a) by action costs which are taken into account for generating optimal plans, i.e., plans that have least total execution cost, and for admissible plans wrt. a given cost bound, i.e., plans whose total execution cost stays within a given limit. As a basis for implementation issues, we have investigated the computational complexity of the major planning tasks in this language, where we have derived complexity results sharply characterizing their computational cost. Furthermore, we have presented a transformation of optimal and admissible planning problems in $\mathcal{K}^c$ to logic programming under the optimal answer set semantics (Buccafurri et al., 1997, 2000), and we have described the $\text{DLV}^{\mathcal{K}}$ prototype implemented on top of the KR tool DLV, which computes this semantics.

As we have shown, $\mathcal{K}^c$ allows for the representation of intricate planning problems. In particular, we have demonstrated this for a variant of the Traveling Salesperson Problem (TSP), which could be conveniently specified in $\mathcal{K}^c$. A strength of $\mathcal{K}^c$ is that, via the underlying language $\mathcal{K}$, states of knowledge, i.e., incomplete states, can be suitably respected in secure plans, i.e., conformant plans which work under all circumstances, including nondeterministic action effects. $\mathcal{K}^c$ is a flexible language which, by exploiting time-dependent action costs, allows for the representation of planning under various optimality criteria such as cheapest plans, shortest plans, and combinations thereof.

Our experiments have shown that various instances of the problems we considered, including realistic instances of the TSP variant, could be decently solved. On the other hand, the current version of DLV$^{\mathcal{K}}$ does not scale to large problem instances in general, and, unsurprisingly, can not compete with high-end planning tools or specialized algorithms for a particular problem such as TSP. We do not see this as a shortcoming, though, since our main goal at this point was to demonstrate the usefulness of the expressive capabilities of our formalism to easily represent non-trivial planning and optimization tasks, which are especially involved from the viewpoint of knowledge representation. In this way, non-trivial instances of such problems of medium size (which one may often encounter) can be solved with little effort.

Several issues remain for further work. As for the implementation, performance improvements may be gained via improvements of the underlying DLV engine, which are subject of current work. Furthermore, alternative, more efficient transformations of $\mathcal{K}^c$ to logic programming might be researched, e.g. ones that involve preprocessing of the planning problem performing means-end analysis to simplify the logic program constructed.

Another issue is further language extensions. For example, a crucial difference between our approach and resource-based approaches is that the former hinges on action costs, while the latter build on fluent values, which is a somewhat different view of the quality of a plan. A possible way to encompass this in our language is to allow that dynamic fluent values contribute to action costs; this needs to be carefully elaborated, though: While for deterministic planning under complete knowledge this extension is straightforward, in non-deterministic domains with incomplete knowledge it would possibly result in ambiguities. Different trajectories of the same plan possibly yield different costs when fluent values contribute to action costs. In favor of an intuitive definition of plan costs and optimality we refrained from this extension at the current state.

A further possible extension are negative action costs, which are useful for modeling producer/consumer relations among actions and resources. Allowing for different priorities among actions, i.e., different cost levels, would increase the flexibility and allow for optimizing different criteria at once. Finally, the duration of actions is an important issue. In the current language, the effects of actions are assumed to materialize in the next state. While by coding techniques, we may express delayed effects over several states in time and/or interleaving actions, constructs in the language would be desirable. Investigating these issues is part of our ongoing and future work.

## Acknowledgments

We are are grateful to Joohyung Lee for his help on using CCALC and to Paul Walser for his useful informations on ILPPLAN. Furthermore, we thank Michael Gelfond for interesting discussions and suggestions, and the anonymous reviewers for their detailed and helpful comments.

This work was supported by FWF (Austrian Science Funds) under the projects P14781 and Z29-N04 and the European Commission under project FET-2001-37004 WASP and IST-2001-33570 INFOMIX.

A preliminary, shorter version of this paper was presented at the 8th European Conference on Logics in Artificial Intelligence (JELIA'02), Cosenza, Italy, September 2002.

## Appendix A. The Language $\mathcal{K}$

This appendix contains, in shortened form, the definition of the language $\mathcal{K}$ and a translation of $\mathcal{K}$ to answer set programs; see (Eiter et al., 2003b, 2003a) for more details and examples.

### A.1 Basic Syntax

We assume $\sigma^{act}$, $\sigma^{fl}$, and $\sigma^{typ}$ disjoint sets of action, fluent and type names, respectively, i.e., predicate symbols of arity $\geq 0$, and disjoint sets $\sigma^{con}$ and $\sigma^{var}$ of constant and variable symbols. Here, $\sigma^{fl}$, $\sigma^{act}$ describe *dynamic knowledge* and $\sigma^{typ}$ describes *static background knowledge*. An *action (resp. fluent, type) atom* is of form $p(t_1, \ldots, t_n)$, where $p \in \sigma^{act}$ (resp. $\sigma^{fl}$, $\sigma^{typ}$) has arity $n$ and $t_1, \ldots, t_n \in \sigma^{con} \cup \sigma^{var}$. An action (resp. fluent, type) literal $l$ is an action (resp. fluent, type) atom $a$ or its negation $\neg a$, where "$\neg$" (alternatively, "–") is the true negation symbol. We define $\neg.l = a$ if $l = \neg a$ and $\neg.l = \neg a$ if $l = a$, where $a$ is an atom. A set $L$ of literals is *consistent*, if $L \cap \neg.L = \emptyset$. Furthermore, $L^+$ (resp. $L^-$) is the set of positive (resp. negative) literals in $L$. The set of all action (resp. fluent, type) literals is denoted as $\mathcal{L}_{act}$ (resp. $\mathcal{L}_{fl}$, $\mathcal{L}_{typ}$). Furthermore, $\mathcal{L}_{fl,typ} = \mathcal{L}_{fl} \cup \mathcal{L}_{typ}$, $\mathcal{L}_{dyn} = \mathcal{L}_{fl} \cup \mathcal{L}^+_{act}$, and $\mathcal{L} = \mathcal{L}_{fl,typ} \cup \mathcal{L}^+_{act}$.

All actions and fluents must be declared using statements as follows.

**Definition A.1 (action, fluent declaration)** *An* action *(resp.* fluent*)* declaration, *is of the form:*

$$p(X_1, \ldots, X_n) \; \texttt{requires} \; t_1, \ldots, t_m \qquad (8)$$

*where $p \in \mathcal{L}^+_{act}$ (resp. $p \in \mathcal{L}^+_{fl}$), $X_1, \ldots, X_n \in \sigma^{var}$ where $n \geq 0$ is the arity of $p$, $t_1, \ldots, t_m \in \mathcal{L}_{typ}$, $m \geq 0$, and every $X_i$ occurs in $t_1, \ldots, t_m$.*

If $m = 0$, the keyword `requires` may be omitted. Causation rules specify dependencies of fluents on other fluents and actions.

**Definition A.2 (causation rule)** *A* causation rule *(*rule*, for short) is an expression of the form*

$$\texttt{caused} \; f \; \texttt{if} \; b_1, \ldots, b_k, \texttt{not} \; b_{k+1}, \ldots, \texttt{not} \; b_l \;\; \texttt{after} \; a_1, \ldots, a_m, \texttt{not} \; a_{m+1}, \ldots, \texttt{not} \; a_n \qquad (9)$$

*where $f \in \mathcal{L}_{fl} \cup \{\texttt{false}\}$, $b_1, \ldots, b_l \in \mathcal{L}_{fl,typ}$, $a_1, \ldots, a_n \mathcal{L}$, $l \geq k \geq 0$, and $n \geq m \geq 0$.*

Rules where $n = 0$ are *static rules*, all others *dynamic rules*. When $l = 0$ (resp. $n = 0$), "`if`" (resp. "`after`") is omitted; if both $l = n = 0$, "`caused`" is optional.

We access parts of a causation rule $r$ by $\mathsf{h}(r) = \{f\}$, $\mathsf{post}^+(r) = \{b_1, \ldots, b_k\}$, $\mathsf{post}^-(r) = \{b_{k+1}, \ldots, b_l\}$, $\mathsf{pre}^+(r) = \{a_1, \ldots, a_m\}$, $\mathsf{pre}^-(r) = \{a_{m+1}, \ldots, a_n\}$, and $\mathsf{lit}(r) = \{f, b_1, \ldots, b_l, a_1, \ldots, a_n\}$. Intuitively, $\mathsf{pre}(r) = \mathsf{pre}^+(r) \cup \mathsf{pre}^-(r)$ (resp. $\mathsf{post}(r) = \mathsf{post}^+(r) \cup \mathsf{post}^-(r)$) accesses the state before (resp. after) some action(s) happen.

Special static rules may be specified for the initial states.

**Definition A.3 (initial state constraint)** *An* initial state constraint *is a static rule of the form (9) preceded by* "`initially`."

The language $\mathcal{K}$ allows conditional execution of actions, where several alternative executability conditions may be specified.

**Definition A.4 (executability condition)** *An* executability condition $e$ *is an expression of the form*

$$\texttt{executable}\ a\ \texttt{if}\ b_1, \ldots, b_k, \texttt{not}\ b_{k+1}, \ldots, \texttt{not}\ b_l \qquad (10)$$

*where* $a \in \mathcal{L}_{act}^+$ *and* $b_1, \ldots, b_l \in \mathcal{L}$, *and* $l \geq k \geq 0$.

If $l = 0$ (i.e., executability is unconditional), "`if`" is skipped. The parts of $e$ are accessed by $\mathsf{h}(e) = \{a\}$, $\mathsf{pre}^+(e) = \{b_1, \ldots, b_k\}$, $\mathsf{pre}^-(e) = \{b_{k+1}, \ldots, b_l\}$, and $\mathsf{lit}(e) = \{a, b_1, \ldots, b_l\}$. Intuitively, $\mathsf{pre}(e) = \mathsf{pre}^+(e) \cup \mathsf{pre}^-(e)$ refers to the state at which some action's suitability is evaluated. The state after action execution is not involved; for convenience, we define $\mathsf{post}^+(e) = \mathsf{post}^-(e) = \emptyset$.

All causal rules and executability conditions must satisfy the following condition, which is similar to safety in logic programs: Each variable in a default-negated type literal must also occur in some literal which is not a default-negated type literal. No safety is requested for variables appearing in other literals. The reason is that variables appearing in fluent and action literals are implicitly safe by the respective type declarations.

*Notation.* For any causal rule, initial state constraint, and executability condition $r$ and $\nu \in \{\mathsf{post}, \mathsf{pre}, \mathsf{b}\}$, we define $\nu(r) = \nu^+(r) \cup \nu^-(r)$, where $\mathsf{b}^s(r) = \mathsf{post}^s(r) \cup \mathsf{pre}^s(r)$.

#### A.1.1 PLANNING DOMAINS AND PLANNING PROBLEMS

**Definition A.5 (action description, planning domain)** *An* action description $\langle D, R\rangle$ *consists of a finite set* $D$ *of action and fluent declarations and a finite set* $R$ *of safe causation rules, safe initial state constraints, and safe executability conditions which do not contain positive cyclic dependencies among actions. A* $\mathcal{K}$ planning domain *is a pair* $PD = \langle \Pi, AD\rangle$, *where* $\Pi$ *is a disjunction-free normal Datalog program (the* background knowledge*) which is safe and has a total well-founded model (cf. (van Gelder, Ross, & Schlipf, 1991))*[8] *and* $AD$ *is an action description. We call* $PD$ positive, *if no default negation occurs in* $AD$.

**Definition A.6 (planning problem)** *A* planning problem $\mathcal{P} = \langle PD, q\rangle$ *is a pair of a planning domain* $PD$ *and a* query $q$, *i.e.,*

$$g_1, \ldots, g_m, \texttt{not}\ g_{m+1}, \ldots, \texttt{not}\ g_n\ ?\ (i) \qquad (11)$$

*where* $g_1, \ldots, g_n \in \mathcal{L}_{fl}$ *are variable-free,* $n \geq m \geq 0$, *and* $i \geq 0$ *denotes the plan length.*

### A.2 Semantics

We start with the preliminary definition of the typed instantiation of a planning domain. This is similar to the grounding of a logic program, with the difference being that only correctly typed fluent and action literals are generated.

Let $PD = \langle \Pi, \langle D, R\rangle\rangle$ be a planning domain, and let $M$ be the (unique) answer set of $\Pi$ (Gelfond & Lifschitz, 1991). Then, $\theta(p(X_1, \ldots, X_n))$ is a *legal action* (resp. *fluent*) *instance* of an action (resp. fluent) declaration $d \in D$ of the form (8), if $\theta$ is a substitution defined over $X_1, \ldots, X_n$ such that $\{\theta(t_1), \ldots, \theta(t_m)\} \subseteq M$. By $\mathcal{L}_{PD}$ we denote the set of all legal action and fluent instances. The instantiation of a planning domain respecting type information is as follows.

8. A total well-founded model, if existing, corresponds to the unique answer set of a datalog program. Allowing for multiple answer sets of $\Pi$ would eventually lead to ambiguities in our language.

**Definition A.7 (typed instantiation)** *For any planning domain* $PD = \langle \Pi, \langle D, R \rangle \rangle$*, its* typed instantiation *is given by* $PD{\downarrow} = \langle \Pi{\downarrow}, \langle D, R{\downarrow} \rangle \rangle$*, where* $\Pi{\downarrow}$ *is the grounding of* $\Pi$ *(over* $\sigma^{con}$*) and* $R{\downarrow} = \{\theta(r) \mid r \in R,\ \theta \in \Theta_r\}$*, where* $\Theta_r$ *is the set of all substitutions* $\theta$ *of the variables in* $r$ *using* $\sigma^{con}$*, such that* $\mathsf{lit}(\theta(r)) \cap \mathcal{L}_{dyn} \subseteq \mathcal{L}_{PD} \cup (\neg.\mathcal{L}_{PD} \cap \mathcal{L}_{fl}^{-})$.

In other words, in $PD{\downarrow}$ we replace $\Pi$ and $R$ by their ground versions, but keep of the latter only rules where the atoms of all fluent and action literals agree with their declarations. We say that a $PD = \langle \Pi, \langle D, R \rangle \rangle$ is *ground*, if $\Pi$ and $R$ are ground, and moreover that it is *well-typed*, if $PD$ and $PD{\downarrow}$ coincide.

### A.2.1 STATES AND TRANSITIONS

**Definition A.8 (state, state transition)** *A* state *w.r.t a planning domain* $PD$ *is any consistent set* $s \subseteq \mathcal{L}_{fl} \cap (\mathsf{lit}(PD) \cup \mathsf{lit}(PD)^{-})$ *of legal fluent instances and their negations. A* state transition *is any tuple* $t = \langle s, A, s' \rangle$ *where* $s, s'$ *are states and* $A \subseteq \mathcal{L}_{act} \cap \mathsf{lit}(PD)$ *is a set of legal action instances in* $PD$.

Observe that a state does not necessarily contain either $f$ or $\neg f$ for each legal instance $f$ of a fluent, and may even be empty ($s = \emptyset$). State transitions are not constrained; this will be done in the definition of *legal state transitions* below. We proceed in analogy to the definition of answer sets (Gelfond & Lifschitz, 1991), considering first positive (i.e., involving a positive planning domain) and then general planning problems.

In what follows, we assume that $PD = \langle \Pi, \langle D, R \rangle \rangle$ is a well-typed ground planning domain and that $M$ is the unique answer set of $\Pi$. For any other $PD$, the respective concepts are defined through its typed grounding $PD{\downarrow}$.

**Definition A.9 (legal initial state)** *A state* $s_0$ *is a* legal initial state *for a positive* $PD$*, if* $s_0$ *is the least set (w.r.t.* $\subseteq$*) such that* $\mathsf{post}(c) \subseteq s_0 \cup M$ *implies* $\mathsf{h}(c) \subseteq s_0$*, for all initial state constraints and static rules* $c \in R$.

For a positive $PD$ and a state $s$, a set $A \subseteq \mathcal{L}_{act}^{+}$ is called *executable action set* w.r.t. $s$, if for each $a \in A$ there exists an executability condition $e \in R$ such that $\mathsf{h}(e) = \{a\}$, $\mathsf{pre}^{+}(e) \cap \mathcal{L}_{fl,typ} \subseteq s \cup M$, $\mathsf{pre}^{+}(e) \cap \mathcal{L}_{act}^{+} \subseteq A$, and $\mathsf{pre}^{-}(e) \cap (\mathcal{L}_{act}^{+} \cup s \cup M) = \emptyset$. Note that this definition allows for modeling dependent actions, i.e. actions which depend on the execution of other actions.

**Definition A.10 (legal state transition)** *Given a positive* $PD$*, a state transition* $t = \langle s, A, s' \rangle$ *is called* legal*, if* $A$ *is an executable action set w.r.t.* $s$ *and* $s'$ *is the minimal consistent set that satisfies all causation rules w.r.t.* $s \cup A \cup M$*. That is, for every causation rule* $r \in R$*, if (i)* $\mathsf{post}(r) \subseteq s' \cup M$*, (ii)* $\mathsf{pre}(r) \cap \mathcal{L}_{fl,typ} \subseteq s \cup M$*, and (iii)* $\mathsf{pre}(r) \cap \mathcal{L}_{act} \subseteq A$ *all hold, then* $\mathsf{h}(r) \neq \{\mathtt{false}\}$ *and* $\mathsf{h}(r) \subseteq s'$.

This is now extended to general a well-typed ground $PD$ containing default negation using a Gelfond-Lifschitz type reduction to a positive planning domain (Gelfond & Lifschitz, 1991).

**Definition A.11 (reduction)** *Let* $PD$ *be a ground and well-typed planning domain, and let* $t = \langle s, A, s' \rangle$ *be a state transition. Then, the* reduction $PD^{t} = \langle \Pi, \langle D, R^{t} \rangle \rangle$ *of* $PD$ *by* $t$ *is the planning domain where* $R^{t}$ *is obtained from* $R$ *by deleting*

1. *each $r \in R$, where either* $\mathsf{post}^-(r) \cap (s' \cup M) \neq \emptyset$ *or* $\mathsf{pre}^-(r) \cap (s \cup A \cup M) \neq \emptyset$*, and*

2. *all default literals* `not` $L$ *($L \in \mathcal{L}$) from the remaining $r \in R$.*

Note that $PD^t$ is positive and ground. We extend further definitions as follows.

**Definition A.12 (legal initial state, executable action set, legal state transition)** *For any planning domain PD, a state $s_0$ is a* legal initial state*, if $s_0$ is a legal initial state for $PD^{\langle\emptyset,\emptyset,s_0\rangle}$; a set $A$ is an* executable action set *w.r.t. a state $s$, if $A$ is executable w.r.t. $s$ in $PD^{\langle s,A,\emptyset\rangle}$; and, a state transition $t = \langle s, A, s'\rangle$ is* legal*, if it is legal in $PD^t$.*

### A.2.2 PLANS

**Definition A.13 (trajectory)** *A sequence of state transitions $T = \langle\langle s_0, A_1, s_1\rangle, \langle s_1, A_2, s_2\rangle, \ldots, \langle s_{n-1}, A_n, s_n\rangle\rangle$, $n \geq 0$, is a* trajectory *for PD, if $s_0$ is a legal initial state of PD and all $\langle s_{i-1}, A_i, s_i\rangle$, $1 \leq i \leq n$, are legal state transitions of PD.*

If $n = 0$, then $T = \langle\rangle$ is empty and has $s_0$ associated explicitly.

**Definition A.14 (optimistic plan)** *A sequence of action sets $\langle A_1, \ldots, A_i\rangle$, $i \geq 0$, is an* optimistic plan *for a planning problem $\mathcal{P} = \langle PD, q\rangle$, if a trajectory $T = \langle\langle s_0, A_1, s_1\rangle, \langle s_1, A_2, s_2\rangle, \ldots, \langle s_{i-1}, A_i, s_i\rangle\rangle$ exists in PD which accomplishes the goal, i.e., $\{g_1, \ldots g_m\} \subseteq s_i$ and $\{g_{m+1}, \ldots, g_n\} \cap s_i = \emptyset$.*

Optimistic plans amount to "plans", "valid plans" etc as defined in the literature. The term "optimistic" should stress the credulous view in this definition, with respect to incomplete fluent information and nondeterministic action effects. In such cases, the execution of an optimistic plan $P$ might fail to reach the goal. We thus resort to secure plans.

**Definition A.15 (secure plans (alias conformant plans))** *An optimistic plan $\langle A_1, \ldots, A_n\rangle$ is a* secure plan*, if for every legal initial state $s_0$ and trajectory $T = \langle\langle s_0, A_1, s_1\rangle, \ldots, \langle s_{j-1}, A_j, s_j\rangle\rangle$ such that $0 \leq j \leq n$, it holds that (i) if $j = n$ then $T$ accomplishes the goal, and (ii) if $j < n$, then $A_{j+1}$ is executable in $s_j$ w.r.t. PD, i.e., some legal transition $\langle s_j, A_{j+1}, s_{j+1}\rangle$ exists.*

Note that plans admit in general the concurrent execution of actions. We call a plan $\langle A_1, \ldots, A_n\rangle$ *sequential* (or *non-concurrent*), if $|A_j| \leq 1$, for all $1 \leq j \leq n$.

## A.3 Macros

$\mathcal{K}$ includes several macros as shorthands for frequently used concepts. Let $\mathtt{a} \in \mathcal{L}^+_{act}$ denote an action atom, $\mathtt{f} \in \mathcal{L}_{fl}$ a fluent literal, `B` a (possibly empty) sequence $b_1, \ldots, b_k$, `not` $b_{k+1}, \ldots,$ `not` $b_l$ where each $b_i \in \mathcal{L}_{fl,typ}, i = 1, \ldots, l$, and `A` a (possibly empty) sequence $a_1, \ldots, a_m$, `not` $a_{m+1}, \ldots,$ `not` $a_n$ where each $a_j \in \mathcal{L}, j = 1, \ldots, n$.

**Inertia** To allow for an easy representation of fluent inertia, $\mathcal{K}$ provides

`inertial f if B after A.` $\Leftrightarrow$ `caused f if not` $\neg$`.f, B after f, A.`

**Defaults** A default value of a fluent can be expressed by the shortcut

`default f.` $\Leftrightarrow$ `caused f if not` $\neg$`.f.`

It is in effect unless some other causation rule provides evidence to the opposite value.

**Totality** For reasoning under incomplete, but total knowledge $\mathcal{K}$ provides (f positive):

| | | |
|---|---|---|
| `total f if B after A.` | $\Leftrightarrow$ | `caused f if not -f, B after A.`<br>`caused -f if not f, B after A.` |

This is is for instance useful to model non-deterministic action effects. For a discussion of the full impact of this statement in modeling planning under incomplete knowledge and non-determinism, we refer to our previous paper on the language $\mathcal{K}$ (Eiter et al., 2003b).

**State Integrity** For integrity constraints that refer to the preceding state, $\mathcal{K}$ provides

`forbidden B after A.` $\Leftrightarrow$ `caused false if B after A.`

**Non-executability** For specifying that some action is *not* executable, $\mathcal{K}$ provides

`nonexecutable a if B.` $\Leftrightarrow$ `caused false after a, B.`

By this definition, `nonexecutable` overrides `executable` in case of conflicts.

**Sequential Plans** To exclude simultaneous execution of actions, $\mathcal{K}$ provides

`noConcurrency.` $\Leftrightarrow$ `caused false after` $\mathtt{a}_1, \mathtt{a}_2$.

where $\mathtt{a}_1$ and $\mathtt{a}_2$ range over all possible actions such that $\mathtt{a}_1, \mathtt{a}_2 \in \mathcal{L}_{PD} \cap \mathcal{L}_{act}$ and $\mathtt{a}_1 \neq \mathtt{a}_2$.

In all macros, "`if B`" (resp. "`after A`") can be omitted, if `B` (resp. `A`) is empty.

## Appendix B. Proofs

**Proof of Theorem 4.4:** *Membership (i):* The problems are in NP resp. NPMV, since if $l$ is polynomial in the size of $\mathcal{P}$, any optimistic plan $P = \langle A_1, \ldots, A_l \rangle$ for $\mathcal{P}$ with a supporting trajectory $T = \langle t_1, \ldots, t_l \rangle$ for $P$ can be guessed and, by Proposition 4.1, verified in polynomial time. Furthermore, $cost_{\mathcal{P}}(P) \leq b$ can be efficiently checked, since $cost_{\mathcal{P}}(P)$ is easily computed (all costs are constants).

*Hardness (i):* $\mathcal{K}$ is a fragment of $\mathcal{K}^c$, and each $\mathcal{K}$ planning problem can be viewed as the problem of deciding the existence of resp. finding an admissible plan wrt. cost 0. As was previously shown (Eiter et al., 2003b), deciding existence of an optimistic plan for a given $\mathcal{K}$ planning problem is NP-hard for fixed plan length $l$; hence, it is also NP-hard for $\mathcal{K}^c$.

We show that finding an optimistic plan is hard for NPMV by a reduction from the well-known SAT problem, cf. (Papadimitriou, 1994), whose instances are CNFs $\phi = c_1 \wedge \cdots \wedge c_k$ of clauses $c_i = L_{i,1} \vee \cdots \vee L_{i,m_i}$, where each $L_{i,j}$ is a classical literal over propositional atoms $X = \{x_1, \ldots, x_n\}$.

Consider the following planning domain $PD_\phi$ for $\phi$:

| | |
|---|---|
| `fluents :` | $x_1$. ... $x_n$. `state0. state1.` |
| `actions :` | $c_1$ `costs 1.` ... $c_k$ `costs 1.` |
| | $ax_1$. ... $ax_n$. |
| `initially :` | `total` $x_1$. ... `total` $x_n$. |
| | `caused state0.` |
| `always :` | `caused state1 after state0.` |
| | `executable` $c_1$ `after` $\neg.L_{1,1}$, ..., $\neg.L_{1,m_1}$. |
| | `forbidden after` $\neg.L_{1,1}$, ..., $\neg.L_{1,m_1}$, `not` $c_1$. |
| | ... |
| | `executable` $c_k$ `after` $\neg.L_{k,1}$, ..., $\neg.L_{k,m_k}$. |
| | `forbidden after` $\neg.L_{k,1}$, ..., $\neg.L_{k,m_k}$, `not` $c_k$. |
| | `executable` $ax_1$ `after` $x_1$. `forbidden after` $x_1$, `not` $ax_1$. |
| | ... |
| | `executable` $ax_n$ `after` $x_n$. `forbidden after` $x_n$, `not` $ax_n$. |

The fluents $x_i$ and `state0` and the `total` statements in the `initially`-section encode the candidate truth assignments. The subsequent statements force $c_j$ to be executed iff the corresponding clause is violated by the truth assignment encoded in the initial state. The final pairs of `executable` and `forbidden` statements force actions $ax_i$ to be executed iff the corresponding fluents $x_i$ hold. This is because it is necessary to directly extract the computed truth assignments from the plan, since we are dealing with a function class. The fluent `state1` identifies the state at time 1.

Consider now the planning problem $\mathcal{P}_\phi = \langle PD_\phi, \mathtt{state1?(1)} \rangle$. Clearly, each optimistic plan $P$ for $\mathcal{P}$ corresponds to a truth assignment $\sigma_P$ of $X$ and vice versa, and $cost_{\mathcal{P}_\phi}(P)$ is the number of clauses violated by $\sigma_P$. Thus, the admissible optimistic plans for $\mathcal{P}_\phi$ wrt. cost 0 correspond 1-1 to the satisfying assignments of $\phi$. Clearly, constructing $\mathcal{P}_\phi$ from $\phi$ is efficiently possible, as is constructing a satisfying truth assignment $\sigma$ from a corresponding plan $P$ (because of the actions $ax_i$). This concludes the hardness proof.

*Membership (ii):* Since the security of each optimistic plan admissible wrt. cost $k$ can be checked, by Proposition 4.1, with a call to a $\Pi_2^P$-oracle, membership in $\Sigma_3^P$ resp. in $\Sigma_3^P\mathrm{MV}$ follows by analogous considerations as in (i) (where no oracle was needed).

*Hardness (ii):* For the decision variant, $\Sigma_3^P$-hardness is again immediately inherited from the $\Sigma_3^P$-completeness of deciding the existence of a secure plan of a problem in the language $\mathcal{K}$, with hardness even for fixed plan length (Eiter et al., 2003b). For the plan computation variant, we give a reduction from the following $\Sigma_3^P\mathrm{MV}$-complete problem: An instance $I$ is an open QBF

$$Q[Z] = \forall X \exists Y \, \Phi[X, Y, Z]$$

where $X = x_1, \ldots, x_l$, $Y = y_1, \ldots, y_m$, and $Z = z_1, \ldots, z_n$, respectively, and $\Phi[X, Y, Z]$ is (w.l.o.g.) a 3CNF formula over $X$, $Y$, and $Z$. The solutions $S(I)$ are all truth assignments over $Z$ for which $Q[Z]$ is satisfied.

Suppose that $\Phi[X, Y, Z] = c_1 \wedge \ldots \wedge c_k$ where $c_i = c_{i,1} \vee c_{i,2} \vee c_{i,3}$. Now consider the following planning domain $PD_{Q[Z]}$ for $Q[Z]$, which is a variant of the planning domain given in the proof of Theorem 5.5 in (Eiter et al., 2003b):

```
fluents :    x1. ... xl. y1. ... ym. z1. ... zn.  state0. state1.
actions :    az1 costs 0. ... azn costs 0.
initially :  total x1. ... total xl.
             caused state0.
always :     caused state1 after state0.
             executable az1. executable az2.... executable azn.
             caused x1 after x1.  caused -x1 after -x1.
             ...
             caused xl after xl.  caused -xl after -xl.
             total y1 after state0. ... total ym after state0.
             caused z1 after az1.  caused -z1 after not az1.
             ...
             caused zn after azn.  caused -zn after not azn.
             forbidden ¬.C1,1, ¬.C1,2, ¬.C1,3 after state0.
             ...
             forbidden ¬.Ck,1, ¬.Ck,2, ¬.Ck,3 after state0.
```

There are $2^{|X|}$ many legal initial states $s^1, \ldots, s^{2^{|X|}}$ for $PD_{Q[Z]}$, which correspond 1-1 to the possible truth assignments to $X$ and all these initial states contain `state0`. Starting from any initial state $s^i$, executing a set of actions represents a truth assignment to the variables in $Z$. Since all actions are always executable, there are $2^{|Z|}$ executable action sets $A_1, \ldots, A_{2^{|Z|}}$, which represent all truth assignments to $Z$.

For each pair $s^i$ and $A_j$ there exist $2^{|Y|}$ many successor state candidates $s^{i,1}, \ldots, s^{i,2^{|Y|}}$, which contain fluents according to the truth assignment to $X$ represented by $s^i$, fluents according to the truth assignment to $Z$ represented by $A_j$, and fluents according to a truth assignment to $Y$, and the fluent `state1`. Of these candidate states, only those satisfying all clauses in $\Phi[X, Y, Z]$ are legal, by virtue of the `forbidden` statements.

It is not hard to see that an optimistic plan of form $P = \langle A_1 \rangle$ (where $A_1 \subseteq \{az_i \mid z_i \in Z\}$) for the goal `state1` exists wrt. $PD_{Q[Z]}$ iff there is an assignment to all variables in $X \cup Y \cup Z$ such that the formula $\Phi[X, Y, Z]$ is satisfied. Furthermore, $P$ is secure iff $A_1$ represents an assignment to the variables in $Z$ such that, regardless of which assignment to the variables in $X$ is chosen (corresponding to a legal initial state $s^i$), there is some assignment to the variables in $Y$ such that all clauses of $\Phi[X, Y, Z]$ are satisfied (i.e., there is at least one state $s^{i,k}$ reachable from $s^i$ by executing $A_1$); any such $s^{i,k}$ contains `state1`. In other words, $P$ is secure iff $\Phi[X, Y, Z]$ is true. Thus, the admissible secure plans of $PD_{Q[Z]}$ wrt. cost 0, correspond 1-1 with the assignments to $Z$ for which $Q[Z]$ is true.

Since $PD_{Q[Z]}$ is constructible from $\Phi[X, Y, Z]$ in polynomial time, it follows that computing a secure plan for $\mathcal{P} = \langle PD_{Q[Z]}, q \rangle$, where $q = \texttt{state1}\ ?\ (\texttt{1})$, is $\Sigma_3^P$MV-hard. □

**Proof of Theorem 4.5:** *Membership (i):* Concerning membership, by performing a binary search on the range $[0, max]$ (where $max$ is an upper bound on the plan costs for a plan of polynomial length $l$ given by $l$ times the sum of all action costs) we can find out the least integer $v$ such that any optimistic plan $P$ for $\mathcal{P}$ which is admissible wrt. cost $v$ exists (if any optimistic plan exists); clearly, we have $cost_{\mathcal{P}}(P) = v$ and $cost^*_{\mathcal{P}} = v$, and thus any such plan $P$ is optimal. Since $max$ is single exponential in the representation size of $\mathcal{P}$, the binary search, and thus computing $cost^*_{\mathcal{P}}$, is, by Theorem 4.4, feasible in polynomial time with an NP oracle. Subsequently, we can construct an optimistic plan $P$ such that $cost_{\mathcal{P}}(P) = cost^*_{\mathcal{P}}$ by extending a partial plan $P_i = \langle A_1, \ldots, A_i \rangle$, $i = 0, \ldots, l-1$ step by step as follows. Let $A = \{a_1, \ldots, a_m\}$ be the set of all legal action instances. We initialize $B_{i+1} := A$ and ask the oracle whether $P_i$ can be completed to an optimistic plan $P = \langle A_1, \ldots, A_l \rangle$ admissible wrt. $cost^*_{\mathcal{P}}$ such that $A_{i+1} \subseteq (B_{i+1} \setminus \{a_1\})$. If the answer is yes, then we update $B_{i+1} := B_{i+1} \setminus \{a_1\}$, else we leave $B_{i+1}$ unchanged. We then repeat this test for $a_j$, $j = 2, 3, \ldots, m$; the resulting $B_{i+1}$ is an action set such that $P_{i+1} = \langle A_1, \ldots, A_i, A_{i+1} \rangle$ where $A_{i+1} = B_{i+1}$ can be completed to an optimistic plan admissible wrt. $cost^*_{\mathcal{P}}$. Thus, $A_{i+1}$ is polynomial-time constructible with an NP oracle.
In summary, we can construct an optimal optimistic plan in polynomial time with an NP oracle. Thus, the problem is in $\mathrm{F}\Delta_2^P$.
*Hardness (i):* We show hardness for plan length $l = 1$ by a reduction from problem MAX WEIGHT SAT (Papadimitriou, 1994), where an instance is a SAT instance $\phi = c_1 \wedge \cdots \wedge c_k$ as in the proof of Theorem 4.4.(i), plus positive integer weights $w_i$, where $i = 1, \ldots, k$. Then, $S(I)$ contains those truth assignments $\sigma$ of $X$ for which $w_{sat}(\sigma) = \sum_{i\,:\,c_i\sigma=\mathrm{true}} w_i$ is maximal.

To that end, we take the planning domain $PD_\phi$ as in the proof of Theorem 4.4 and modify the cost of $c_i$ to $w_i$, for $i = 1, \ldots, k$, thus constructing a new planning domain $PD_I$. Consider now the planning problem $\mathcal{P}_I = \langle PD_I, \texttt{state1?(1)} \rangle$. Since the actions $c_j$ are the only actions with nonzero cost, any plan (corresponding to a truth assignment $\sigma$) will be associated with the sum of weights of violated clauses, $w_{vio}(\sigma) = (\sum_{i=1}^k w_i) - w_{sat}(\sigma)$. Since $\sum_{i=1}^k w_i$ is constant for $I$, minimizing $w_{vio}(\sigma)$ is equivalent to maximizing $w_{sat}(\sigma)$. Hence, there is a one-to-one correspondence between optimal optimistic plans of $\mathcal{P}_I$ (for which $w_{vio}(\sigma)$ is minimal) and maximal truth assignments for $I$. Furthermore, computing $\mathcal{P}_I$ from $I$ and extracting a MAX-WEIGHT SAT solution from an optimal plan $P$ is efficiently possible. This proves $\mathrm{F}\Delta_2^P$-hardness.
*Membership (ii):* The proof is similar to the membership proof of (i), but uses an oracle which asks for completion of a partial secure plan $P_i = \langle A_1, \ldots, A_i \rangle$ to a secure plan $P = \langle A_1, \ldots, A_l \rangle$ such that $A_{i+1} \subseteq (B_{i+1} \setminus \{a_j\})$ and $P$ is admissible wrt. $cost^*_{\mathcal{P}}$, rather than of a partial optimistic plan. This oracle is, as easily seen, in $\Sigma_3^P$. Thus, computing an optimal secure plan is in $\mathrm{F}\Delta_4^P$.
*Hardness (ii):* We show hardness by a reduction from the following problem, which is $\mathrm{F}\Delta_4^P$-complete (cf. (Krentel, 1992)): Given an open QBF $Q[Z] = \forall X \exists Y \Phi[X, Y, Z]$ like in the proof of Theorem 4.4.(ii), compute the lexicographically first truth assignment of $Z$ for which $Q[Z]$ is satisfied.

This can be accomplished by changing the cost of each action $az_i$ in $PD_{Q[Z]}$ from $0$ to $2^{n-i}$, $i = 1, \ldots, n$. Let $PD'[Q[Z]]$ be the resulting planning domain. Since the cost of $az_i$ (i.e., assigning $z_i$ the value true) is greater than the sum of the costs of all $az_j$ for $i + 1 \leq j \leq n$, an optimal secure plan for the planning problem $\langle PD'[Q[Z]], \texttt{state1}\,?\,(\texttt{1}) \rangle$ amounts to the lexicographically first truth assignment for $Z$ such that $Q[Z]$ is satisfied. Thus, $\mathrm{F}\Delta_4^P$-hardness of the problem follows.
□

**Proof of Theorem 6.1:** We prove the result by applying the well-known Splitting Set Theorem for logic programs (Lifschitz & Turner, 1994). This theorem applies to logic programs $\pi$ that can be split into two parts such that one of them, the "bottom" part, does not refer to predicates defined in the "top" part at all. The answer sets of the "bottom" part can then be extended to the answer sets of the whole program by looking at the remaining ("top") rules. Informally, a splitting set of $\pi$ is a set $U$ of ground literals defining the "bottom" part $b_U(\pi)$ of a program. Each answer set $S_b$ of $b_U(\pi)$ can then be used to reduce the remaining rules $\pi \setminus b_U(\pi)$ to a program $e_U(\pi \setminus b_U(\pi), S_b)$ involving only classical literals which do not occur in $b_U(\pi)$, by evaluating the literals from $b_U(\pi)$ wrt. $S_b$. For each answer set $S_e$ of $e_U(\pi \setminus b_U(\pi), S_b)$, the set $S = S_b \cup S_e$ then is an answer set of the original program.

Disregarding weak constraints, we can split the program $lp^w(\mathcal{P})$ into a bottom part consisting of $lp(\mathcal{P}_{nc})$, where $\mathcal{P}_{nc}$ is $\mathcal{P}$ with the cost information stripped off, and a top part containing the remaining rules; we then derive the correspondence between optimistic plans for $\mathcal{P}$ and answer sets of $lp^w(\mathcal{P})$ from a similar correspondence result for $lp(\mathcal{P}_{nc})$ (Eiter et al., 2003a).

In detail, Theorem 3.1 in (Eiter et al., 2003a) states for any $\mathcal{K}$-planning problem $\mathcal{P}$ a correspondence between the answer sets $S$ of $lp(\mathcal{P})$ and supporting trajectories $T$ of optimistic plans $P = \langle A_1, \ldots, A_l \rangle$ as in items (i) and (ii), with costs discarded. Thus, any answer set $S'$ of $lp(\mathcal{P}_{nc})$ corresponds to some trajectory $T'$ of an optimistic plan $P'$ for $\mathcal{P}_{nc}$ and vice versa.

In what follows, when talking about $lp(\mathcal{P}_{nc})$ and $lp^w(\mathcal{P})$, we mean the respective grounded logic programs. $lp^w(\mathcal{P})$ augments $lp(\mathcal{P}_{nc})$ by rules (4) and weak constraints (5). Let now $U = lit(lp(\mathcal{P}_{nc}))$ be the set of all literals occurring in $lp(\mathcal{P}_{nc})$. Clearly, $U$ splits $lp^w(\mathcal{P})$ as defined in (Lifschitz & Turner, 1994), where we disregard weak constraints in $lp^w(\mathcal{P})$, since the rules of form (4) introduce only new head literals. Consequently, we get $b_U(lp^w(\mathcal{P})) = lp(\mathcal{P}_{nc})$. Then, for any answer set $S'$ of $lp(\mathcal{P}_{nc})$, each rule in $e_U(lp^w(\mathcal{P}) \setminus b_U(lp^w(\mathcal{P})), S')$ is of the form

$$cost_a(x_1, \ldots, x_n, t, c) \text{ :- } Body.$$

From the fact that all these rules are positive, we can conclude that with respect to the split by $U$, any answer set $S'$ of $lp(\mathcal{P}_{nc})$ induces a unique answer set $S \supseteq S'$ of $lp^w(\mathcal{P})$. Therefore, modulo costs, a correspondence between supporting trajectories $T$ and candidate answer sets $S$ as claimed follows directly from Theorem 3.1 in (Eiter et al., 2003a).

It remains to prove that $cost_{\mathcal{P}}(P) = \mathsf{cost}_{lp^w(\mathcal{P})}(S)$ holds for all candidate answer sets $S$ corresponding to an optimistic plan $P = \langle A_1, \ldots, A_l \rangle$ for $\mathcal{P}$. By the correspondence shown above, any action $p(x_1, \ldots, x_n) \in A_j$ corresponds to exactly one atom $p(x_1, \ldots, x_n, j-1) \in A_j^S$, $j \in \{1, \ldots, l\}$. Therefore, if $p(x_1, \ldots, x_n)$ is declared with a non-empty `cost` part, by (4) and well-definedness, modulo $x_1, \ldots, x_n$, there is exactly one fact $cost_p(x_1, \ldots, x_n, j-1, c)$ in the model of $e_U(lp^w(\mathcal{P}) \setminus b_U(lp^w(\mathcal{P})), S)$.

Furthermore, by definition of (4), we have that $c = cost_j(p(x_1, \ldots, x_n))$, i.e., the cost of action instance $p(x_1, \ldots, x_n)$ at time $j$. Consequently, the violation value of the weak constraint $wc$ of form (5) for $p$ in $lp^w(\mathcal{P})$ is $\mathsf{cost}_{wc}(S) = \sum_{j=1}^{l} \sum_{p(x_1,\ldots,x_n) \in A_j} cost_j(p(x_1, \ldots, x_n))$. Since all violation values stem from weak constraints (5), in total we have $\mathsf{cost}_{lp^w(\mathcal{P})}(S) = cost_{\mathcal{P}}(P)$. This proves the result. □

## References

Blum, A. L., & Furst, M. L. (1997). Fast Planning Through Planning Graph Analysis. *Artificial Intelligence*, *90*, 281–300.

Bonet, B., & Geffner, H. (2000). Planning with Incomplete Information as Heuristic Search in Belief Space. In Chien, S., Kambhampati, S., & Knoblock, C. A. (Eds.), *Proceedings of the Fifth International Conference on Artificial Intelligence Planning and Scheduling (AIPS'00)*, pp. 52–61, Breckenridge, Colorado, USA.

Bryant, R. E. (1986). Graph-based algorithms for boolean function manipulation. *IEEE Transactions on Computers*, *C-35*(8), 677–691.

Buccafurri, F., Leone, N., & Rullo, P. (1997). Strong and Weak Constraints in Disjunctive Datalog. In Dix, J., Furbach, U., & Nerode, A. (Eds.), *Proceedings of the 4th International Conference on Logic Programming and Non-Monotonic Reasoning (LPNMR'97)*, No. 1265 in Lecture Notes in AI (LNAI), pp. 2–17, Dagstuhl, Germany. Springer Verlag.

Buccafurri, F., Leone, N., & Rullo, P. (2000). Enhancing Disjunctive Datalog by Constraints. *IEEE Transactions on Knowledge and Data Engineering*, *12*(5), 845–860.

Cimatti, A., & Roveri, M. (2000). Conformant Planning via Symbolic Model Checking. *Journal of Artificial Intelligence Research*, *13*, 305–338.

Dantsin, E., Eiter, T., Gottlob, G., & Voronkov, A. (2001). Complexity and Expressive Power of Logic Programming. *ACM Computing Surveys*, *33*(3), 374–425.

Dimopoulos, Y., Nebel, B., & Koehler, J. (1997). Encoding Planning Problems in Nonmonotonic Logic Programs. In *Proceedings of the European Conference on Planning 1997 (ECP-97)*, pp. 169–181. Springer Verlag.

Eiter, T., Faber, W., Leone, N., & Pfeifer, G. (2000a). Declarative Problem-Solving Using the DLV System. In Minker, J. (Ed.), *Logic-Based Artificial Intelligence*, pp. 79–103. Kluwer Academic Publishers.

Eiter, T., Faber, W., Leone, N., Pfeifer, G., & Polleres, A. (2000b). Planning under Incomplete Knowledge. In Lloyd, J., Dahl, V., Furbach, U., Kerber, M., Lau, K.-K., Palamidessi, C., Pereira, L. M., Sagiv, Y., & Stuckey, P. J. (Eds.), *Computational Logic - CL 2000, First International Conference, Proceedings*, No. 1861 in Lecture Notes in AI (LNAI), pp. 807–821, London, UK. Springer Verlag.

Eiter, T., Faber, W., Leone, N., Pfeifer, G., & Polleres, A. (2002a). Answer Set Planning under Action Costs. In Flesca, S., Greco, S., Ianni, G., & Leone, N. (Eds.), *Proceedings of the 8th European Conference on Artificial Intelligence (JELIA)*, No. 2424 in Lecture Notes in Computer Science, pp. 186–197.

Eiter, T., Faber, W., Leone, N., Pfeifer, G., & Polleres, A. (2002b). Answer Set Planning under Action Costs. Tech. rep. INFSYS RR-1843-02-13, Institut für Informationssysteme, Technische Universität Wien.

Eiter, T., Faber, W., Leone, N., Pfeifer, G., & Polleres, A. (2003a). A Logic Programming Approach to Knowledge-State Planning, II: the DLV$^{\mathcal{K}}$ System. *Artificial Intelligence*, *144*(1–2), 157–211.

Eiter, T., Faber, W., Leone, N., Pfeifer, G., & Polleres, A. (2003b). A Logic Programming Approach to Knowledge-State Planning: Semantics and Complexity. To appear in ACM Transactions on Computational Logic.

Ephrati, E., Pollack, M. E., & Mihlstein, M. (1996). A Cost-directed Planner: Preliminary Report. In *Proceedings of the Thirteenth National Conference on Artificial Intelligence (AAAI-96)*, pp. 1223 – 1228. AAAI Press.

Erdem, E. (1999). Applications of Logic Programming to Planning: Computational Experiments. Unpublished draft. `http://www.cs.utexas.edu/users/esra/papers.html`.

Faber, W., & Pfeifer, G. (since 1996). DLV homepage.. `http://www.dlvsystem.com/`.

Ferraris, P., & Giunchiglia, E. (2000). Planning as Satisfiability in Nondeterministic Domains. In *Proceedings of the Seventeenth National Conference on Artificial Intelligence (AAAI'00), July 30 – August 3, 2000, Austin, Texas USA*, pp. 748–753. AAAI Press / The MIT Press.

Fourer, R., Gay, D. M., & Kernighan, B. W. (1993). *AMPL: A Modeling Language for Mathematical Programming*. Duxbury Press.

Gelfond, M., & Lifschitz, V. (1991). Classical Negation in Logic Programs and Disjunctive Databases. *New Generation Computing*, *9*, 365–385.

Ghallab, M., Howe, A., Knoblock, C., McDermott, D., Ram, A., Veloso, M., Weld, D., & Wilkins, D. (1998). PDDL — The Planning Domain Definition language. Tech. rep., Yale Center for Computational Vision and Control. Available at http://www.cs.yale.edu/pub/mcdermott/software/pddl.tar.gz.

Giunchiglia, E. (2000). Planning as Satisfiability with Expressive Action Languages: Concurrency, Constraints and Nondeterminism. In Cohn, A. G., Giunchiglia, F., & Selman, B. (Eds.), *Proceedings of the Seventh International Conference on Principles of Knowledge Representation and Reasoning (KR 2000), April 12-15, Breckenridge, Colorado, USA*, pp. 657–666. Morgan Kaufmann.

Giunchiglia, E., Kartha, G. N., & Lifschitz, V. (1997). Representing Action: Indeterminacy and Ramifications. *Artificial Intelligence, 95*, 409–443.

Giunchiglia, E., & Lifschitz, V. (1998). An Action Language Based on Causal Explanation: Preliminary Report. In *Proceedings of the Fifteenth National Conference on Artificial Intelligence (AAAI '98)*, pp. 623–630.

Haslum, P., & Geffner, H. (2000). Admissible Heuristics for Optimal Planning. In Chien, S., Kambhampati, S., & Knoblock, C. A. (Eds.), *Proceedings of the Fifth International Conference on Artificial Intelligence Planning and Scheduling (AIPS'00)*, pp. 140–149, Breckenridge, Colorado, USA. AAAI Press.

Kautz, H., & Walser, J. P. (1999). State-space planning by integer optimization. In *Proceedings of the 16th National Conference on Artificial Intelligence (AAAI-99)*, pp. 526–533.

Koehler, J. (1998). Planning Under Resource Constraints. In *Proceedings of the 13th European Conference on Artificial Intelligence (ECAI'98)*, pp. 489–493.

Krentel, M. (1992). Generalizations of Opt P to the Polynomial Hierarchy. *Theoretical Computer Science*, *97*(2), 183–198.

Lee, J., & Lifschitz, V. (2001). Additive Fluents. In Provetti, A., & Cao, S. T. (Eds.), *Proceedings AAAI 2001 Spring Symposium on Answer Set Programming: Towards Efficient and Scalable Knowledge Representation and Reasoning*, pp. 116–123, Stanford, CA. AAAI Press.

Lifschitz, V., & Turner, H. (1994). Splitting a Logic Program. In Van Hentenryck, P. (Ed.), *Proceedings of the 11th International Conference on Logic Programming (ICLP'94)*, pp. 23–37, Santa Margherita Ligure, Italy. MIT Press.

Lifschitz, V., & Turner, H. (1999). Representing Transition Systems by Logic Programs. In Gelfond, M., Leone, N., & Pfeifer, G. (Eds.), *Proceedings of the 5th International Conference on Logic Programming and Nonmonotonic Reasoning (LPNMR'99)*, No. 1730 in Lecture Notes in AI (LNAI), pp. 92–106, El Paso, Texas, USA. Springer Verlag.

Lifschitz, V. (1996). Foundations of Logic Programming. In Brewka, G. (Ed.), *Principles of Knowledge Representation*, pp. 69–127. CSLI Publications, Stanford.

Lifschitz, V. (1999a). Action Languages, Answer Sets and Planning. In Apt, K., Marek, V. W., Truszczyński, M., & Warren, D. S. (Eds.), *The Logic Programming Paradigm – A 25-Year Perspective*, pp. 357–373. Springer Verlag.

Lifschitz, V. (1999b). Answer Set Planning. In Schreye, D. D. (Ed.), *Proceedings of the 16th International Conference on Logic Programming (ICLP'99)*, pp. 23–37, Las Cruces, New Mexico, USA. The MIT Press.

McCain, N. (1999). The Causal Calculator Homepage.. `http://www.cs.utexas.edu/users/tag/cc/`.

McCain, N., & Turner, H. (1997). Causal Theories of Actions and Change. In *Proceedings of the 15th National Conference on Artificial Intelligence (AAAI-97)*, pp. 460–465.

McCain, N., & Turner, H. (1998). Satisfiability Planning with Causal Theories. In Cohn, A. G., Schubert, L., & Shapiro, S. C. (Eds.), *Proceedings Sixth International Conference on Principles of Knowledge Representation and Reasoning (KR'98)*, pp. 212–223. Morgan Kaufmann Publishers.

Moskewicz, M. W., Madigan, C. F., Zhao, Y., Zhang, L., & Malik, S. (2001). Chaff: Engineering an Efficient SAT Solver. In *Proceedings of the 38th Design Automation Conference, DAC 2001, Las Vegas, NV, USA, June 18-22, 2001*, pp. 530–535. ACM.

Nareyek, A. (2001). Beyond the Plan-Length Criterion. In *Local Search for Planning and Scheduling, ECAI 2000 Workshop*, Vol. 2148 of *Lecture Notes in Computer Science*, pp. 55–78. Springer.

Niemelä, I. (1998). Logic Programs with Stable Model Semantics as a Constraint Programming Paradigm. In Niemelä, I., & Schaub, T. (Eds.), *Proceedings of the Workshop on Computational Aspects of Nonmonotonic Reasoning*, pp. 72–79, Trento, Italy.

Papadimitriou, C. H. (1994). *Computational Complexity*. Addison-Wesley.

Pednault, E. P. D. (1989). Exploring the Middle Ground between STRIPS and the Situation Calculus. In *Proceedings of the 1st International Conference on Principles of Knowledge Representation and Reasoning (KR'89)*, pp. 324–332, Toronto, Canada. Morgan Kaufmann Publishers, Inc.

Refanidis, I., & Vlahavas, I. (2001). A Framework for Multi-Criteria Plan Evaluation in Heuristic State-Space Planning. In *IJCAI-01 Workshop on Planning with Resources*.

Selman, A. L. (1994). A Taxonomy of Complexity Classes of Functions. *Journal of Computer and System Sciences*, *48*(2), 357–381.

Simons, P., Niemelä, I., & Soininen, T. (2002). Extending and Implementing the Stable Model Semantics. *Artificial Intelligence*, *138*, 181–234.

Smith, D. E., & Weld, D. S. (1998). Conformant Graphplan. In *Proceedings of the Fifteenth National Conference on Artificial Intelligence, (AAAI'98)*, pp. 889–896. AAAI Press / The MIT Press.

Son, T. C., & Pontelli, E. (2002). Reasoning About Actions in Prioritized Default Theory. In Flesca, S., Greco, S., Ianni, G., & Leone, N. (Eds.), *Proceedings of the 8th European Conference on Artificial Intelligence (JELIA)*, No. 2424 in Lecture Notes in Computer Science, pp. 369–381.

Subrahmanian, V., & Zaniolo, C. (1995). Relating Stable Models and AI Planning Domains. In Sterling, L. (Ed.), *Proceedings of the $12^{th}$ International Conference on Logic Programming*, pp. 233–247, Tokyo, Japan. MIT Press.

van Gelder, A., Ross, K., & Schlipf, J. (1991). The Well-Founded Semantics for General Logic Programs. *Journal of the ACM*, *38*(3), 620–650.

Weld, D. S., Anderson, C. R., & Smith, D. E. (1998). Extending Graphplan to Handle Uncertainty & Sensing Actions. In *Proceedings of the Fifteenth National Conference on Artificial Intelligence, (AAAI'98)*, pp. 897–904. AAAI Press / The MIT Press.

Williams, M., & Hanks, S. (1994). Optimal Planning with a Goal-Directed Utility Model. In Hammond, K. J. (Ed.), *Proceedings of the Second International Conference on Artificial Intelligence Planning Systems (AIPS-94)*, pp. 176–181. AAAI Press.